# Video-Based Markerless Motion Capture for Clinical and Rehabilitation Biomechanics: A PRISMA-ScR Scoping Review of Validated Architectures, Clinical Readiness, and Emerging Methods

Florian Delaplace[1], Elodie Piche[1 2], Frédéric Chorin[2], Raphael Zory[1 3]

[1] LAMHESS, Université Côte d'Azur, Nice, France · [2] Plateforme Fragilité, Centre Hospitalier Universitaire de Nice, Clinique Gériatrique de Soins Ambulatoires, Nice, France · [3] Institut Universitaire de France (IUF), France

*Corresponding author: Florian Delaplace, florian.delaplace@unice.fr*

## Abstract

**Background.** Video-based markerless motion capture promises movement analysis without the cost, space and skin-marker constraints of optoelectronic systems, with particular potential for clinical and rehabilitation settings. Whether validated pipelines yet deliver clinically acceptable biomechanics, and how they relate to the underlying computer-vision research, remains unclear.

**Methods.** We conducted a scoping review following the PRISMA extension for Scoping Reviews, with a registered protocol and searches of PubMed, Scopus and IEEE Xplore (January 2015 to February 2026; the computer-vision scan was updated to July 2026). A dual-tier design paired a primary corpus of validated biomechanical studies with a complementary, curated and deliberately non-exhaustive corpus of emerging computer-vision work, used qualitatively. We charted study characteristics, pipeline architecture, validation methods and joint-angle accuracy.

**Results.** We included 117 studies, most published from 2024 onward and conducted on healthy adults walking in a laboratory. Pipelines formed five architectural families across monocular and multi-camera modalities; most reported raw joint angles without biomechanical refinement. Sagittal lower-limb agreement clustered around 5 to 6°, generally short of clinical acceptability, while out-of-plane kinematics, kinetics, and pathological or older populations were rarely validated. Emerging computer-vision building blocks (foundation-model mesh recovery, differentiable inverse kinematics, video-based kinetics) were almost absent from validated studies.

**Conclusions.** Video-based markerless capture is not yet interchangeable with marker-based systems for clinical joint kinematics, and it remains barely validated where rehabilitation needs it most: older and pathological populations, out-of-plane kinematics, and kinetics. Mapping this evidence gap onto emerging computer-vision advances, we propose hypothesis-generating design guidelines, not a validated method, to steer the next generation of pipelines toward accessible, clinically meaningful movement analysis.

## 1. Background

Three-dimensional motion capture has been a cornerstone of biomechanical research and clinical gait analysis for over three decades. The optoelectronic, marker-based motion capture (MBMC) systems (Vicon, Qualisys and OptiTrack) remain the de facto gold standard for joint kinematics measurement, with documented sub-millimetre spatial accuracy and sub-degree joint angle precision under controlled laboratory conditions [1]. Yet their adoption beyond high-budget academic and clinical research labs has been constrained by three barriers: capital cost (tens of thousands to hundreds of thousands of euros per system), space (dedicated capture volumes with controlled lighting), and operational overhead (marker placement, soft-tissue artefact, post-processing labour). The accuracy of even a well-set-up optoelectronic system is bounded less by its hardware than by soft-tissue artefact, the movement of skin markers relative to bone [2-3]; removing skin markers is therefore part of the appeal of video-based approaches.

In parallel, the past decade has seen rapid growth in computer vision (CV) techniques for human pose estimation from RGB video alone, the so-called "markerless motion capture" (MMC) paradigm [4]. From foundational 2D pose estimators (OpenPose [5], BlazePose [6], HRNet [7]) to multi-view triangulation pipelines (Pose2Sim [8], OpenCap [9]) and, more recently, monocular 3D recovery via deep neural networks (MotionBERT [10], BioPose [11], OpenCap Monocular [12]), the field has produced a broad set of algorithmic approaches. Potential gains include equipment cost reduction by one to three orders of magnitude, deployment in clinical or home settings, and workflows compatible with both research and routine clinical use.

Yet this technological promise has not yet been translated into a coherent clinical reality. The current literature on MMC for biomechanics exhibits substantial heterogeneity: distinct algorithmic families coexist without consensus on which architecture is preferable for a given clinical or research question; reported accuracy metrics span an order of magnitude across studies; clinically-relevant joint planes (frontal, transverse) are systematically under-reported relative to the sagittal plane; and validation across pathological populations remains scarce. These shortfalls carry direct clinical weight. The frontal- and transverse-plane deviations that are least reported, such as knee valgus, hip internal rotation, pelvic obliquity, circumduction and scissoring, are precisely the features that define common neurological gait disorders [13], so a pipeline that resolves only the sagittal plane to within five to six degrees can miss the very signs a rehabilitation clinician needs to track. Moreover, much of the cutting-edge computer vision research appears almost exclusively in preprint repositories (arXiv) and computer-vision conference proceedings. Integrating such newly published algorithmic building blocks into a pipeline, validating it against a reference standard, and publishing the result all take time, so a non-negligible delay is expected before the most recent computer-vision advances reach the validated biomedical literature. This review examines that expectation.

The breadth, heterogeneity, and rapid evolution of this evidence base render a meta-analytic synthesis impractical and arguably misleading: the variability in study design, reporting metrics, gold-standard equipment, and target populations across studies precludes inferential meta-analysis (weighted effect-size estimation); we therefore summarise the evidence descriptively, with medians, ranges and proportions, rather than pooling effect sizes. A scoping review, conducted following the PRISMA-ScR (Preferred Reporting Items for Systematic Reviews and Meta-Analyses extension for Scoping Reviews) and the Joanna Briggs Institute (JBI) methodological framework, is therefore a more appropriate instrument to map this evidence.

Prior reviews of markerless biomechanics have largely catalogued architectures and reported accuracy [4, 14]; what remains missing is twofold: an account of how the clinically validated field connects to the fast-moving computer-vision frontier that will shape its next generation, and a step beyond pure description toward concrete, evidence-anchored design directions. We address both by deliberately bridging two literatures that rarely meet, clinical biomechanics and computer vision. Beyond the conventional scoping review, we propose a dual-tier design. The MACRO level constitutes the formal PRISMA screening of biomedical databases (PubMed, Scopus, IEEE Xplore) and yields the corpus of studies that have empirically validated an MMC pipeline against an external biomechanical gold standard. The MICRO level constitutes a complementary, curated corpus from two sources: published records screened out of the systematic search because they addressed only a single pipeline component rather than a complete validated pipeline, and a targeted scan of recent arXiv preprints and computer-vision conference proceedings, used to surface the technological building blocks that underpin the next generation of MMC pipelines. The MICRO is non-exhaustive by design and is used qualitatively, to identify and characterise emerging technical families rather than to estimate volumes, but explicitly compensates for the structural delay between preprint-era CV research and its eventual indexation in biomedical databases.

This scoping review pursues two research questions.

RQ1 (MACRO): What are the architectures, validation outcomes, and clinical applicability of currently-published video-based markerless motion capture pipelines for biomechanical analysis?

RQ2 (MICRO): What are the emerging technical building blocks in the computer vision literature (2024-2026) that, characterised through a deliberately non-systematic horizon scan, could shape the next generation of MMC pipelines but have not yet transitioned into peer-reviewed biomedical publications?

Synthesising both, we then distil a set of evidence-based design guidelines for clinically deployable markerless pipelines, across two deployment modalities (monocular, single-camera and clinic- or field-deployable; and multi-camera, laboratory-grade), combining the empirical lessons from validated MACRO studies with the technical state of the art from MICRO. This third strand is deliberately forward-looking: it is presented as a perspective and research agenda, offering hypothesis-generating directions anchored to the gaps this review documents, not a validated method.

## 2. Methods

### 2.1 Study design and protocol

This scoping review was conducted in accordance with the PRISMA-ScR guidelines [15] and the JBI methodological framework for scoping reviews [16]. The review protocol was prospectively registered with the Open Science Framework (OSF, doi: 10.17605/OSF.IO/TBYPC); the completed PRISMA-ScR checklist is provided in Additional file 1. A scoping review design was selected over a systematic review for two reasons: first, the substantial heterogeneity in study design, validation methodology and reported metrics across the included literature precludes meaningful meta-analytic pooling; second, our explicit objective is to map and characterise a rapidly-evolving evidence landscape, including its emerging technical fringes, rather than to estimate a pooled effect.

### 2.2 Information sources and search strategy

We searched three biomedical and engineering databases (PubMed, Scopus and IEEE Xplore) for studies published between January 2015 and February 2026, with the last database search executed on 4 February 2026. The search strategy combined controlled vocabulary (MeSH for PubMed) and free-text terms covering markerless motion capture, video-based pose estimation, and biomechanical validation. Full search queries are reproduced in Additional file 2 (Appendix A).

To complement the formal biomedical search, we paired the systematic review with an algorithmic horizon scan: a second, deliberately non-exhaustive corpus (the MICRO corpus) of emerging computer-vision work. It drew on two sources: screening records from the systematic search that addressed only a single pipeline component, and a targeted scan of recent computer-vision literature, namely arXiv (primarily the cs.CV category) and the proceedings of major vision and learning venues (CVPR, ICCV, ECCV, WACV, NeurIPS), bounded to 2024-2026 to capture methods not yet indexed in biomedical databases. Its construction is detailed in Section 2.6. The MICRO corpus is used qualitatively, to surface technical building blocks and emerging families, not to estimate their prevalence. The computer-vision scan was run separately from the systematic search and updated to July 2026, later than the February 2026 biomedical search cut-off, so a small number of recent preprints post-date the MACRO search.

## 2.3 Eligibility criteria

Eligibility was articulated using the Population-Concept-Context (PCC) framework, with explicit inclusion and exclusion criteria for the MACRO corpus.

**Inclusion criteria**

Studies were included if they: (i) evaluated a markerless motion capture pipeline based on video input (single or multi-camera); (ii) reported quantitative validation against an external biomechanical reference standard (Vicon, Qualisys, OptiTrack, BTS/MIQUS, Motion Analysis, Nokov, or comparable optoelectronic system; force plates for kinetics; or, in a minority of cases, instrumented IMU systems); (iii) reported empirical joint-angle, spatiotemporal, or kinetic accuracy metrics (RMSE, MAE, correlation coefficients, or ICC); (iv) focused on biomechanical motion analysis rather than general pose detection or activity recognition.

**Exclusion criteria**

Eight pre-specified exclusion criteria (A-H) were applied:

A. Black-box system: commercial or proprietary pipelines whose pose-estimation model is not identified or disclosed, for example Theia3D, Captury and KinaTrax. Such systems were excluded because their pose-estimation model cannot be characterised or reproduced, even though Theia3D in particular is among the most clinically deployed markerless systems; eleven studies were excluded here.

B. No modern computer vision: manual digitisation or classical (non-deep-learning) CV only.

C. Depth-sensor channel: studies using a depth or RGB-D modality (e.g. Kinect) as the primary three-dimensional mechanism, excluded because this review concerns RGB-video pipelines, although such devices remain common in rehabilitation clinics.

D. Review without primary data.

E. Not markerless: physical body markers used as input to the pipeline.

F. Manual gold standard only: validation against Kinovea, goniometer, video manual annotation, or comparable manual reference.

G. No biomechanical output: only intermediate metrics (mean-per-joint-position error, MPJPE, which scores keypoint position rather than joint angle) reported, no joint angles, spatiotemporal parameters, or kinetics.

H. No external gold standard: only internal test-retest reliability reported, with no comparison to an independent reference system.

Studies were considered for the MICRO corpus regardless of validation status, provided they addressed at least one pipeline processing component (2D pose estimation, 3D lifting, triangulation, mesh recovery, inverse kinematics, gait event detection, temporal modelling, physics-informed methods, or signal processing).

## 2.4 Study selection process

Both title-and-abstract screening and full-text assessment against the eligibility criteria were performed independently by two reviewers using Covidence systematic review software. Conflicts at either stage were resolved by a third, senior reviewer. Full-text exclusions were coded into broad reason categories (wrong technology, validation, outcome, document type, population, or report not retrieved), while studies that passed full text but failed the pre-specified A-H criteria at data-extraction verification were excluded at that later stage (Fig. 1).

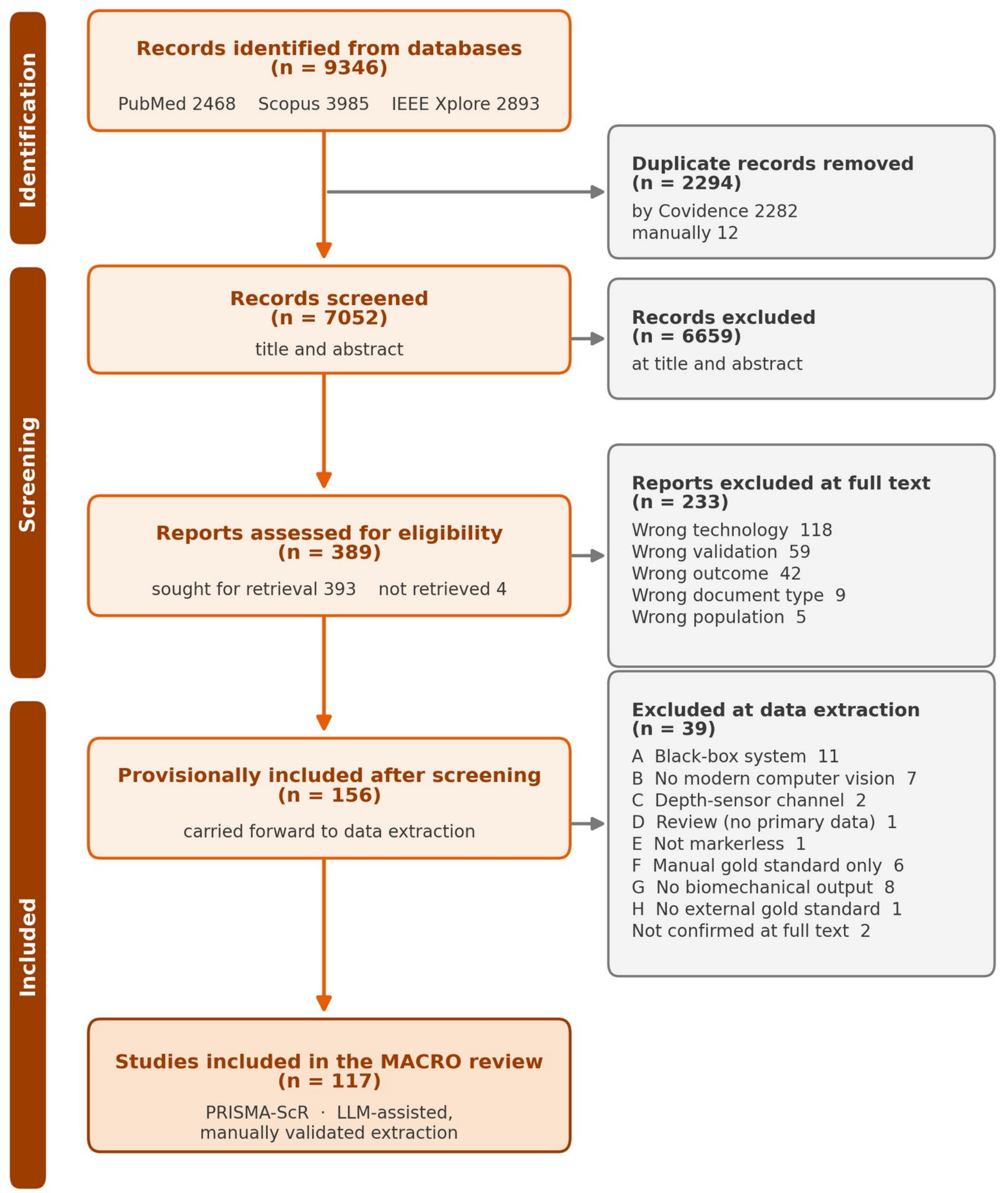

*Figure 1. PRISMA-ScR flow diagram of study identification, screening and inclusion.*

## 2.5 Data extraction

We developed a structured extraction template in Microsoft Excel comprising 7 sheets and 90+ data fields, organised across six domains:

General characteristics (study ID, full title, first author, year, paper type, sample size, participant demographics, target population, motor task, clothing).

Video input specifications (number of cameras, positioning, camera type, resolution, acquisition frequency, calibration procedure, synchronisation, environment).

Computer vision processing (2D pose algorithm, backbone architecture, training regime, training dataset, 3D reconstruction method, pipeline name, occlusion handling, filtering, marker augmentation, real-time capability).

Biomechanical processing (biomechanical software, kinematics method, skeletal model, scaling procedure, kinetics/forces computation).

Global and spatiotemporal validation (gold standard system, validated outcome categories, reported metrics, global kinematic error and correlation, spatiotemporal errors, authors' conclusion).

Joint-level validation (per-joint × per-plane RMSE and correlation: pelvis, hip, knee, ankle × sagittal, frontal, transverse).

A seventh sheet synthesised the per-study architectural classification into the 5-family taxonomy used in the analysis.

**LLM-assisted data extraction**

Data extraction was performed by the authors, with large language models used as an assistive, error-surfacing step rather than as the source of record. The first author defined the structured template (about 90 fields across the 117 studies). To speed first-pass population of the template from heterogeneous sources (free text, tables, figures, scanned PDFs), each study was processed by a large language model (Claude Opus 4.7). The table was then independently re-extracted and cross-checked by a second model from a different vendor (Google Gemini), and every divergence between the two was adjudicated by the authors against the source PDF. Before adjudication, the two models diverged on 156 of approximately 10,500 fields (about 90 fields across 117 studies), a raw agreement of approximately 98.5%. A single chance-corrected agreement coefficient was not computed because the template mixes numeric, categorical and free-text fields. Across all verification passes, 385 cell-level corrections were applied (more than the 156 model disagreements, because some errors were shared by both models and surfaced only on checking against the source PDF), these being predominantly numeric transcription errors and per-joint or per-plane misalignments, with a smaller share of categorical misclassifications of architecture or gold standard. Two instances in which a model reported a value from a cited prior study rather than the paper itself were also identified and corrected. As a final internal consistency check, the first author manually re-extracted a random 20% of the included studies (selected with a computer random-number generator) directly from the source documents; all fields in this sample matched the table exactly (identical numeric values and categorical labels). This final pass was not independent of the primary extraction; independence rests on the two-model, cross-vendor adjudication described above. The values are therefore machine-assisted but human-verified, and the authors are responsible for the extracted dataset.

## 2.6 Data synthesis

Synthesis proceeded across three hierarchical tiers.

**Descriptive analysis**

The MACRO corpus was first characterised in terms of temporal trends, population characteristics, sample sizes, motor tasks, and capture environments.

Architectural mapping. The 117 MACRO studies were classified into a five-family architectural taxonomy (2D Monocular, Multi-view Triangulation, Monocular Lifting 2D→3D, Body Mesh Recovery/SMPL, End-to-End Direct), with five non-exclusive overlay variants (+IK/MSK, +Marker Augmentation, +Temporal Deep Learning, +Physics, Geometric Only). For each family we report the distribution of technical components (backbone, pose estimator, biomechanical software) and the validation outcomes per joint and anatomical plane. Joint-angle errors were summarised against McGinley's error bands (2° or less acceptable, 2 to 5° reasonable, above 5° likely to mislead) as a clinically relevant benchmark.

Technological deconstruction. The MICRO corpus was built in two steps. First, records identified during database screening that described only a single pipeline component, rather than a complete pipeline validated against a gold standard (exclusion criteria F and G), were retained and grouped into 17 pipeline-step blocks (3D lifting, mesh recovery, 2D pose, physics-informed methods, triangulation, gait events, depth, IK solvers, filtering, motion prediction, marker augmentation, tracking, musculoskeletal modelling, temporal smoothing, foundation models, kinetics, and calibration; Fig. 6, Section 3.4). Second, we searched further within each block, targeting recent arXiv preprints and computer-vision conference papers (2024-2026) to capture emerging state-of-the-art methods absent from the biomedical screening; each emerging method was then traced back to the MACRO architectural family it could upgrade. The MICRO corpus is thus curated and deliberately non-exhaustive. The arXiv (cs.CV) and conference scan was run up to July 2026; the complete record list, the source of each entry, and the per-block counts (Additional file 6) are provided in Additional file 5. This MICRO scan sits outside PRISMA-ScR, which governs the systematic MACRO corpus only. The full construction protocol and per-block reproducible queries are provided in Additional file 7.

Finally, a MACRO-MICRO bridge analysis articulated, for each of the five architectural families, the corresponding technical building blocks in the MICRO corpus and the principal points of divergence between the validated biomedical literature and the emerging CV state-of-the-art (Section 3.5).

# 3. Results

## 3.1 Description of the included corpus

From 9,346 records identified across the three databases, 2,294 duplicates were removed and 7,052 titles and abstracts were screened; 6,659 were excluded and 393 full-text reports were assessed, of which 237 were excluded at full text and 39 more at data-extraction verification against criteria A-H, yielding the 117 included studies (Fig. 1). The 117 included studies span January 2015 to February 2026; per-study characteristics are provided in Additional file 3. Publication accelerated recently: about six in ten appeared from 2024 onward (Fig. 2a).

The evidence base is narrow. Participants are overwhelmingly healthy adults. Clinical or pathological populations (stroke, Parkinson's disease, cerebral palsy, spinal-cord injury) appear in a minority of studies (about 19%), and paediatric or older-adult cohorts are rarer still (Fig. 2b). Gait is the dominant motor task, present in more than half of the corpus; running, sit-to-stand, jump-landing and squatting account for most of the remainder (Fig. 2c). Sample sizes are small, with a median of 17 participants (Fig. 2d). The demographic skew and small samples limit the generalisability of individual validation studies.

Validation almost always uses an optoelectronic marker-based system as the reference (about 83% of studies; Fig. 2e). Instrumented walkways, force plates and public datasets cover the rest. Data acquisition remains laboratory-based: 100 of 117 studies took place in a laboratory and 16 in clinical movement-analysis rooms, with home or outdoor capture almost absent (one study and none). Validated evidence for analysis outside the optical laboratory is therefore lacking.

The technical configuration confirms a research-prototype stage (Table 1). Acquisition uses a single camera in 45% of studies and two cameras in 29%, with machine-vision cameras (40%) or consumer smartphones (26%). Calibration is absent in just under half of studies (53) and otherwise uses a checkerboard. Two-dimensional pose relies on the OpenPose family in about half of studies, a count that includes OpenCap deployments whose 2D backend (OpenPose by default, or optionally HRNet) is usually not reported. MediaPipe and BlazePose [6] come next. Three-dimensional recovery splits almost evenly between multi-view triangulation and purely two-dimensional analysis. Monocular lifting, mesh fitting and end-to-end regression each hold a small share. Biomechanical processing uses custom code (Python or Matlab) in 63% of studies, and OpenSim-based modelling in just over a quarter of studies (33), of which 13 used the turnkey OpenCap smartphone pipeline and 20 used OpenSim directly. Real-time operation is reported in 8 of 117 studies (7%). Offline post-processing is the norm.

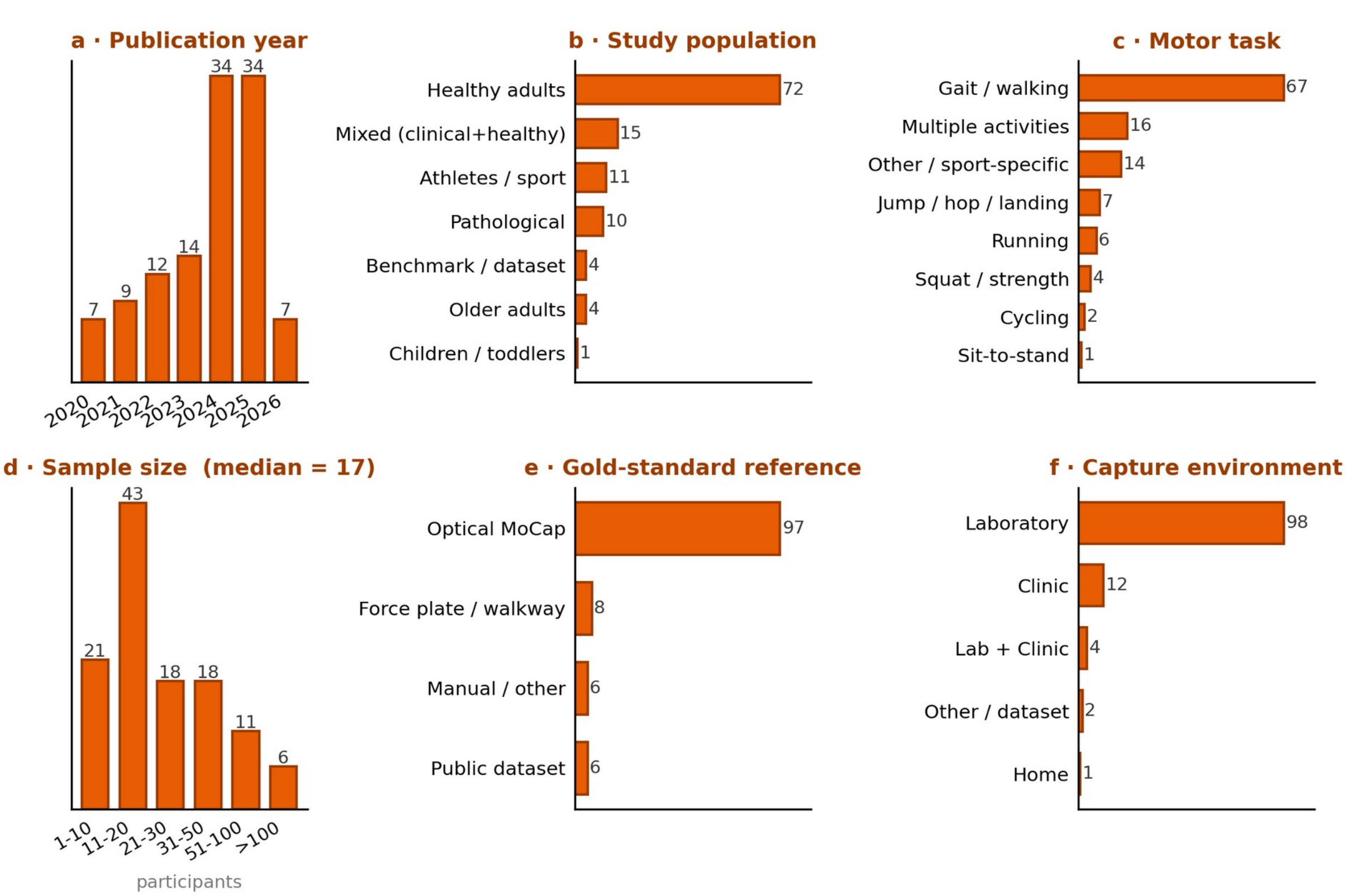

***Figure 2. Overview of the included corpus (n = 117): a, publication year; b, study population; c, motor task; d, sample size; e, gold-standard reference; f, capture environment.***

**Table 1. Technical configuration of the included corpus (n = 117).**

| **Variable / category** | n (%) |
|---|---|
| Number of cameras | |
| 1 | 53 (45%) |
| 2 | 34 (29%) |
| 3-4 | 13 (11%) |
| 5-9 | 11 (9%) |
| >=10 | 4 (3%) |
| Benchmark / varies | 2 (2%) |
| Camera type | |
| Industrial / machine-vision | 47 (40%) |
| Smartphone | 30 (26%) |
| Webcam | 14 (12%) |
| Action cam / DSLR / RGB | 16 (14%) |
| Other / not reported | 5 (4%) |
| Tablet | 5 (4%) |
| Calibration | |
| None (monocular) | 53 (45%) |
| Checkerboard | 38 (32%) |
| Other / not specified | 20 (17%) |
| Bundle adjustment / dot-matrix | 3 (3%) |
| Wand | 3 (3%) |
| 2D pose estimator | |
| OpenPose / OpenCap family | 57 (49%) |
| MediaPipe / BlazePose | 17 (15%) |
| Other named DL detectors (AlphaPose, YOLO, PifPaf, HRNet, MMPose, Detectron2, ViTPose…) | 17 (15%) |
| Custom / unspecified DL | 7 (6%) |
| DeepLabCut | 6 (5%) |
| None / end-to-end | 5 (4%) |
| Proprietary / SDK | 5 (4%) |
| Multi-estimator benchmark | 3 (3%) |
| Architecture (3D recovery) | |
| 2D Monocular (no explicit 3D) | 47 (40%) |
| Multi-view triangulation | 44 (38%) |
| Monocular lifting (2D->3D) | 19 (16%) |
| Body mesh / SMPL | 4 (3%) |
| End-to-end / direct | 3 (3%) |
| Biomechanical software | |
| Custom code (Python/Matlab) | 74 (63%) |
| OpenSim-based modelling | 33 (28%) |
| via OpenCap (smartphone app) | 13 (11%) |
| OpenSim directly | 20 (17%) |
| None (2D angles) | 8 (7%) |
| Other MSK (Visual3D, CusToM, AnyBody) | 2 (2%) |
| Real-time capability | |
| No (post-process) | 107 (92%) |
| Yes / near real-time | 8 (7%) |
| Not specified | 2 (2%) |

## 3.2 Core pipeline architectures

Each pipeline belongs to one of five mutually exclusive architectural families, grouped by deployment modality (Fig. 3). Three terms are kept distinct throughout: modality (reconstruction from a single view, monocular, or from multiple synchronised views), family (one of the five architectures), and overlay (an optional refinement block any family may add). The monocular modality gathers four families that differ mainly in how they reach three dimensions: 2D Monocular computes joint angles directly from two-dimensional keypoints (47 studies, 40%) [17-63]; Monocular Lifting regresses 3D pose from those detections (19, 16%) [64-81, 86]; Body Mesh Recovery fits a parametric whole-body mesh rather than a sparse skeleton (4, 3%) [11, 82-84], its exemplar BioPose [11] being a peer-reviewed WACV 2025 paper retrieved through the IEEE Xplore search; and End-to-End Direct regresses joint angles straight from images (3, 3%) [85, 87-88]. The multi-camera modality is dominated by Multi-view Triangulation, which reconstructs 3D joint positions by intersecting rays from several calibrated cameras (44, 38%) [8-9, 89-130]. The two modalities differ in volume (73 vs 44 studies) and, more importantly, in capability (Section 3.3). Modality is not the same as physical camera count. Many single-view pipelines used extra cameras only for comparison or calibration, and one End-to-End study used multi-view input, so families are assigned by architecture, not by camera count (Table 1).

2D Monocular and Triangulation are well established and still growing (27 and 30 studies from 2024 onward). Monocular Lifting emerged after 2021 and is expanding. Mesh recovery is the newest entrant: every SMPL-based study appeared in 2024 or later. End-to-end regression appeared briefly and has not been sustained. The recent rise in triangulation studies follows the release of turnkey multi-view tools such as Pose2Sim and OpenCap.

An architecture is a shared pipeline skeleton onto which optional refinement blocks are added, not a fixed method. Five non-exclusive overlays were coded: inverse kinematics with musculoskeletal modelling, virtual-marker augmentation, deep-learning temporal smoothing, physics-based optimisation, and a kinetics output (ground-reaction forces or joint moments). Most pipelines add little refinement. Just under two-thirds (73 of 117) use no overlay and report raw geometric angles, and only 28 combine two or more. Refinement concentrates in the multi-view family. Inverse kinematics with musculoskeletal modelling is the most common overlay (36 studies, 31%) and is applied by 27 of the 44 triangulation studies, which also hold almost all virtual-marker-augmentation pipelines (the OpenCap pattern). In contrast, 46 of the 47 2D Monocular studies remain geometric-only, and OpenSim-based use (including OpenCap) is almost confined to triangulation. Serious musculoskeletal modelling therefore sits on the multi-camera side, while the monocular side, which has the greatest deployment potential, mostly stops at raw two-dimensional angles.

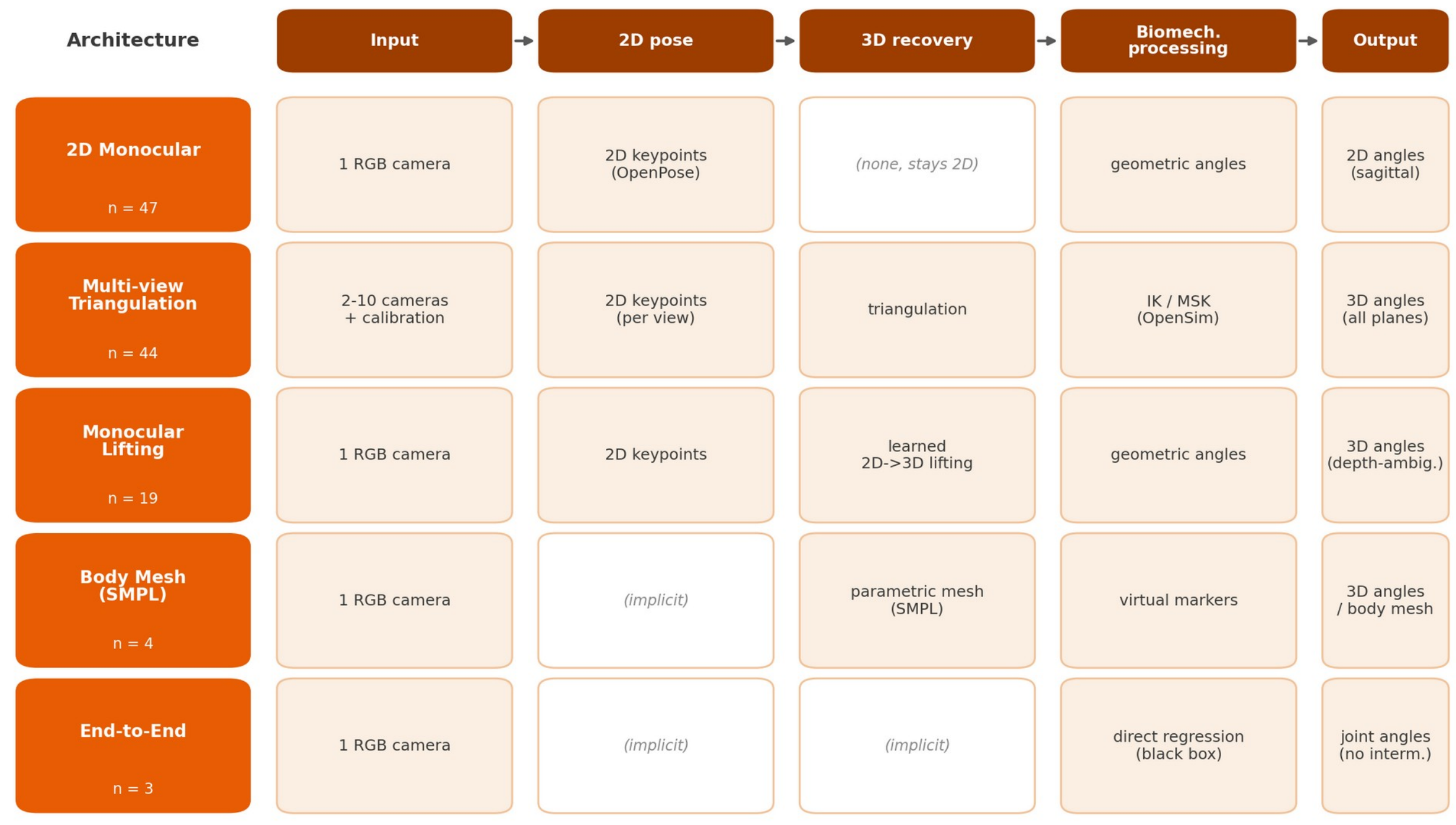

***Figure 3. The five markerless motion-capture architectures and their shared pipeline stages.***

## 3.3 Metrological validation

### 3.3.1 Reference standards

About 83% of studies validated against an optoelectronic marker-based system, most often Vicon, Qualisys, OptiTrack, BTS or Nokov. A minority used an instrumented walkway (GAITRite, Zeno/PKMAS), force plates, a public motion-capture dataset, or, rarely, planar radiography. These reference systems are not themselves error-free: they carry soft-tissue artefact, the movement of skin markers relative to bone [2], which sets a ceiling on the agreement any markerless system can show against them. Optoelectronic capture nonetheless remains the de facto reference standard for biomechanics. A recent markerless study even found its own inconsistencies to be at least as large as that artefact [117], so the two paradigms are not cleanly separable on accuracy.

### 3.3.2 Joint-angle accuracy and the clinical threshold

Sagittal-plane lower-limb agreement clustered around a median absolute error of roughly 5 to 6° across hip, knee, ankle and pelvis, with all four joints within about a degree of one another (Table 2). Inter-study spread was wide (interquartile ranges of roughly 3 to 8°). We benchmarked agreement against the error bands proposed by McGinley et al. for three-dimensional gait kinematics, where errors of 2° or less are widely regarded as acceptable, errors of 2 to 5° as reasonable but requiring consideration during interpretation, and errors above 5° as likely to mislead clinical interpretation [131]. These are marker-based reliability figures. Because the reference systems carry their own repeatability, which can itself approach 5° outside the sagittal plane, agreement near this level is partly bounded by the reference's own noise and is better read as not yet within marker-based reliability than as clinically unusable. Acceptable error is not a single constant, since it depends on the clinical question and on the size of the deviation being measured, so we report a band rather than one cut-off. Across sagittal lower-limb estimates only about 6% fall within the 2° acceptable band and about 37% within the 2 to 5° reasonable band, while, considered separately, about half of sagittal lower-limb estimates exceed 5° under root-mean-square error alone (median 5.0°) and about half under mean absolute error alone (median 5.0°). Pooling all reported metrics gives a similar figure (about 57%). This is the level above which McGinley et al. consider errors likely to mislead clinical interpretation (Fig. 5). The pattern, not the exact rate, is the finding. Most published pipelines exceed the 5° concern level even where they perform best. In rehabilitation terms, an error of five to six degrees is of the same order as the kinematic change a clinician seeks to track during recovery or treatment titration, for example the gradual reduction of a crouch posture in cerebral palsy or of stiff-knee and circumduction patterns after stroke, so an instrument at this error level can mask genuine clinical change.

#### 3.3.3 Accuracy by architecture

We charted sagittal error by architecture family (Fig. 4). The two best-represented families, 2D Monocular (n = 47) and Multi-view Triangulation (n = 44), performed similarly, with median lower-limb errors of roughly 4 to 6°. Monocular Lifting (n = 19) performed less well, with a median lower-limb error around 6 to 7° and the widest inter-study spread. This is consistent with depth ambiguity: a single camera cannot fully resolve motion toward or away from it, so regressing three-dimensional pose from one view is intrinsically harder than triangulating several calibrated views. Body Mesh Recovery and End-to-End Direct were each represented by four or fewer studies, too few to compare, and are charted descriptively only. No family reached the 2° acceptable band at the median, so the clinical-accuracy gap is a property of the field as a whole, not of a single weak architecture.

#### 3.3.4 Plane coverage and the sagittal reporting bias

The main limitation is partly what is reported and partly what can be measured at all. Out-of-plane kinematics come almost only from the multi-view family, for two distinct reasons: the 2D Monocular family cannot recover out-of-plane rotation at all, as it produces no three-dimensional estimate, whereas the monocular three-dimensional families and triangulation can recover it but mostly do not report it. In the architecture × joint × plane map (Additional file 4), frontal and transverse cells are populated mainly for triangulation (for example hip frontal n = 13, hip transverse n = 10), while every monocular family fills a single sagittal column. The reporting imbalance is large. For the knee, 65 studies reported sagittal error against 5 frontal and 3 transverse; the hip (57 / 17 / 12) and ankle (49 / 4 / 1) follow the same pattern. Where frontal and transverse values exist they are both scarcer and larger (for example ankle frontal about 9°). The planes most relevant to several clinical questions, including knee valgus and varus, hip rotation and foot progression, are the least validated.

### 3.3.5 Spatiotemporal and kinetic validation

Spatiotemporal parameters (walking speed, step length, step time) were validated in about a third of studies (34%) and agreed more closely than joint angles, reflecting their lower sensitivity to keypoint noise. Kinetic validation against force plates, for ground-reaction forces or joint moments, was reported in only nine studies. The field validates mainly sagittal lower-limb kinematics. Out-of-plane kinematics and kinetics remain largely unexamined, as Sections 3.4 and 3.5 develop.

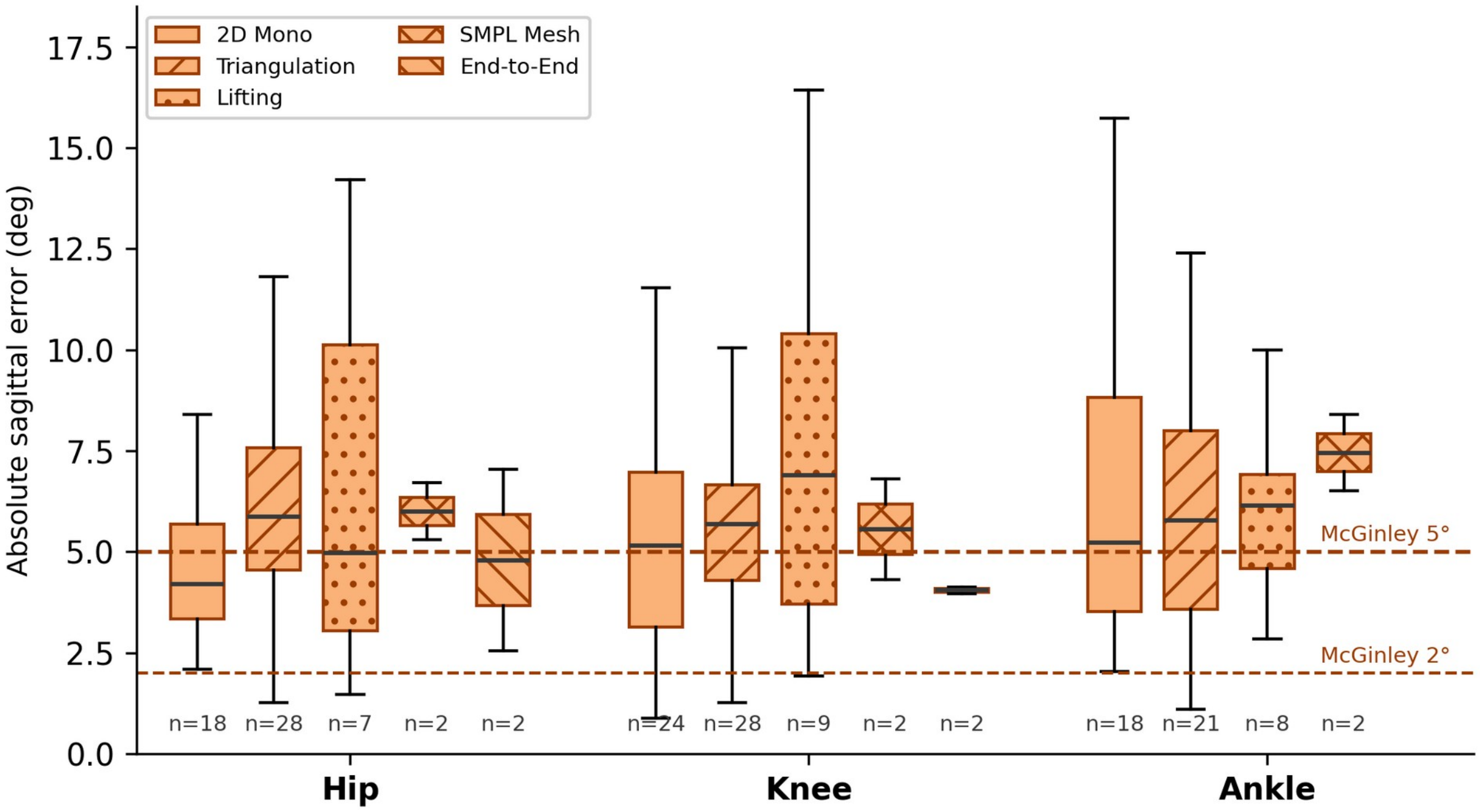

*Figure 4. Sagittal-plane lower-limb joint error by core architecture.*

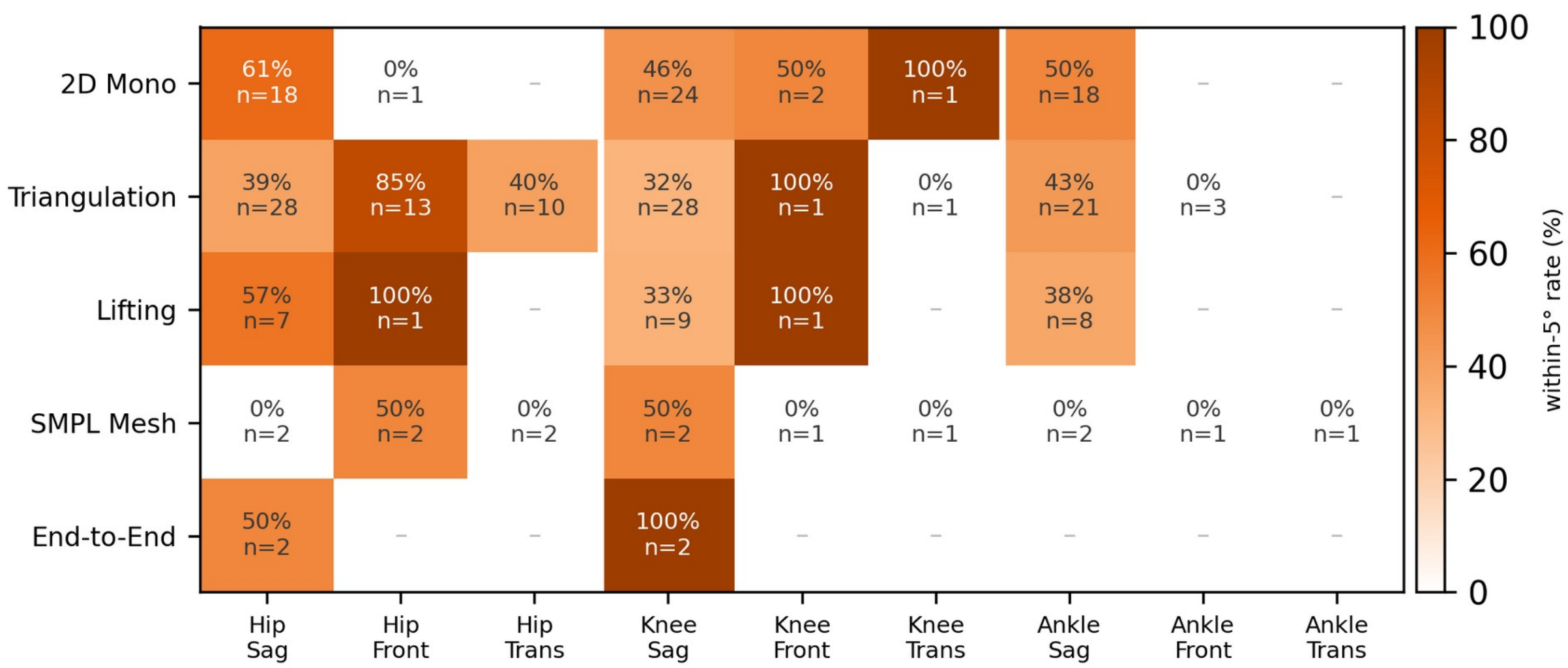

***Figure 5. Proportion of studies within the 5° error level (McGinley acceptable-to-reasonable range), by architecture, joint and plane. Each cell is computed over the studies contributing that architecture, joint and plane.***

**Table 2. Joint-angle validation by anatomical plane (absolute agreement vs optoelectronic reference).**

| **Joint** | **Plane** | **n studies** | **Median \|error\| (°)** | **IQR (°)** | ≤ 5° (%) |
|---|---|---|---|---|---|
| Pelvis | Sagittal | 5 | 5.2 | 3.4-6.1 | 40 |
| Pelvis | Frontal | 3 | 4.0 | 2.0-4.9 | — |
| Pelvis | Transverse | 3 | 3.1 | 2.0-3.3 | — |
| Hip | Sagittal | 57 | 5.5 | 3.6-7.1 | 47 |
| Hip | Frontal | 17 | 4.3 | 3.0-5.2 | 76 |
| Hip | Transverse | 12 | 5.7 | 4.4-7.0 | 33 |
| Knee | Sagittal | 65 | 5.6 | 3.7-7.0 | 40 |
| Knee | Frontal | 5 | 4.8 | 3.1-12.8 | 60 |
| Knee | Transverse | 3 | 5.9 | 2.4-9.5 | — |
| Ankle | Sagittal | 49 | 5.9 | 3.6-8.4 | 43 |
| Ankle | Frontal | 4 | 9.0 | 5.9-10.3 | — |
| Ankle | Transverse | 1 | 11.1 | n/a | — |

*Note: marked under-reporting outside the sagittal plane (e.g. knee: 65 sagittal vs 5 frontal vs 3 transverse). The final column reports the proportion within the 5° error level (McGinley acceptable-to-reasonable range; see Section 3.3.2). Cells based on fewer than five studies are indicative only. Reported agreement metrics pool both root-mean-square error and mean absolute error owing to widespread reporting heterogeneity; per-metric breakdowns are provided in Additional file 3. Restricting to the two principal metrics leaves this shortfall unchanged: about half of sagittal lower-limb estimates exceed 5° under root-mean-square error alone (median 5.0°) and under mean absolute error alone (median 5.0°).*

## 3.4 Technological deconstruction (MICRO)

To identify the computer-vision building blocks behind the next pipeline generation, we assembled the complementary MICRO corpus of 572 arXiv and conference records and grouped them into 17 pipeline-step blocks. Because this corpus is curated rather than systematic (Methods), Figure 6 characterises each block by its relative level of activity (low, moderate or high), not by a raw count, and we draw no MACRO-to-MICRO ratio.

One qualitative signal is worth noting, with caution. The records we screened cluster in 2024 and 2025, and the foundation-model block is almost entirely post-2024. Because the corpus was curated and seeded by backward citation, this recency partly reflects our sampling rather than the field, so we read it only as a coarse, qualitative hint of the delay between new computer-vision methods and their uptake in validated biomedical work, not as a measured interval. The clearer evidence for this gap is the near-absence of these emerging blocks from the validated corpus (Section 3.5).

Read as a toolbox mapped onto the pipeline stages (Fig. 7; the full list of methods, the stage each targets, its contribution and its maturity are given in Additional file 5), the state of the art offers a candidate upgrade for nearly every block [132-150]. At the input stage, monocular metric-depth and 3D-geometry models and target-free multi-camera self-calibration are maturing [174-176], and world-grounding through camera-motion, gravity or learned field-of-view cues helps place the subject metrically from a single camera, a problem that remains only partly solved [133, 177]. For 2D pose, real-time and transformer backbones [151-153] are replacing the OpenPose generation that dominates the MACRO corpus. For 3D recovery several families coexist: monocular lifting [10, 154]; foundation-model mesh recovery on the SMPL family [155-157, 172], which reframes the pipeline around a dense parametric mesh and gives access to anatomical markers [158]; whole-body mesh-to-angle inverse kinematics [171]; multi-person and multi-view mesh fitting [169, 173]; and classical triangulation to OpenSim [8]. For the biomechanical bridge, virtual-marker augmentation [159] and differentiable or neural inverse kinematics [160-162], with scaling and inverse dynamics [168], are emerging. The kinetics block estimates ground-reaction forces and joint moments downstream of pose with physics-informed or learned models [163], targeting an output the MACRO literature rarely validates. Above reconstruction, an analysis layer with multimodal biomechanics foundation models [146] and per-degree-of-freedom uncertainty quantification [164] is beginning to appear, addressing the reliability requirement of clinical deployment. Several near-complete pipelines compose these blocks rather than introducing a new one: BioPose [11], OpenCap Monocular [12] and the Portable Biomechanics Laboratory [170], the last reporting agreement its authors describe as clinically acceptable from handheld video.

|  |  |  |
|---|---|---|
| Input / sensing | Calibration | Low |
| | Depth | High |
| 2D pose | 2D pose | High |
| 3D recovery | 3D lifting | High |
| | Mesh recovery | High |
| | Triangulation | High |
| | Foundation models | Low |
| | Motion prediction | Moderate |
| | Tracking | Low |
| Biomechanical bridge | IK solvers | Moderate |
| | Marker augmentation | Low |
| | Musculoskeletal modelling | Low |
| | Physics-informed | Moderate |
| Temporal / signal | Filtering | Moderate |
| | Temporal smoothing | Low |
| Output / analysis | Kinetics / GRF | Moderate |
| | Gait events | High |

***Figure 6. Qualitative activity of the curated MICRO pool across the 17 pipeline-step blocks. Shading denotes the relative level of representation (low, moderate, high) within a non-systematic, curated pool; it reflects screening and curation effort, not field size or biomechanical importance, and should not be read as a measured quantity.***

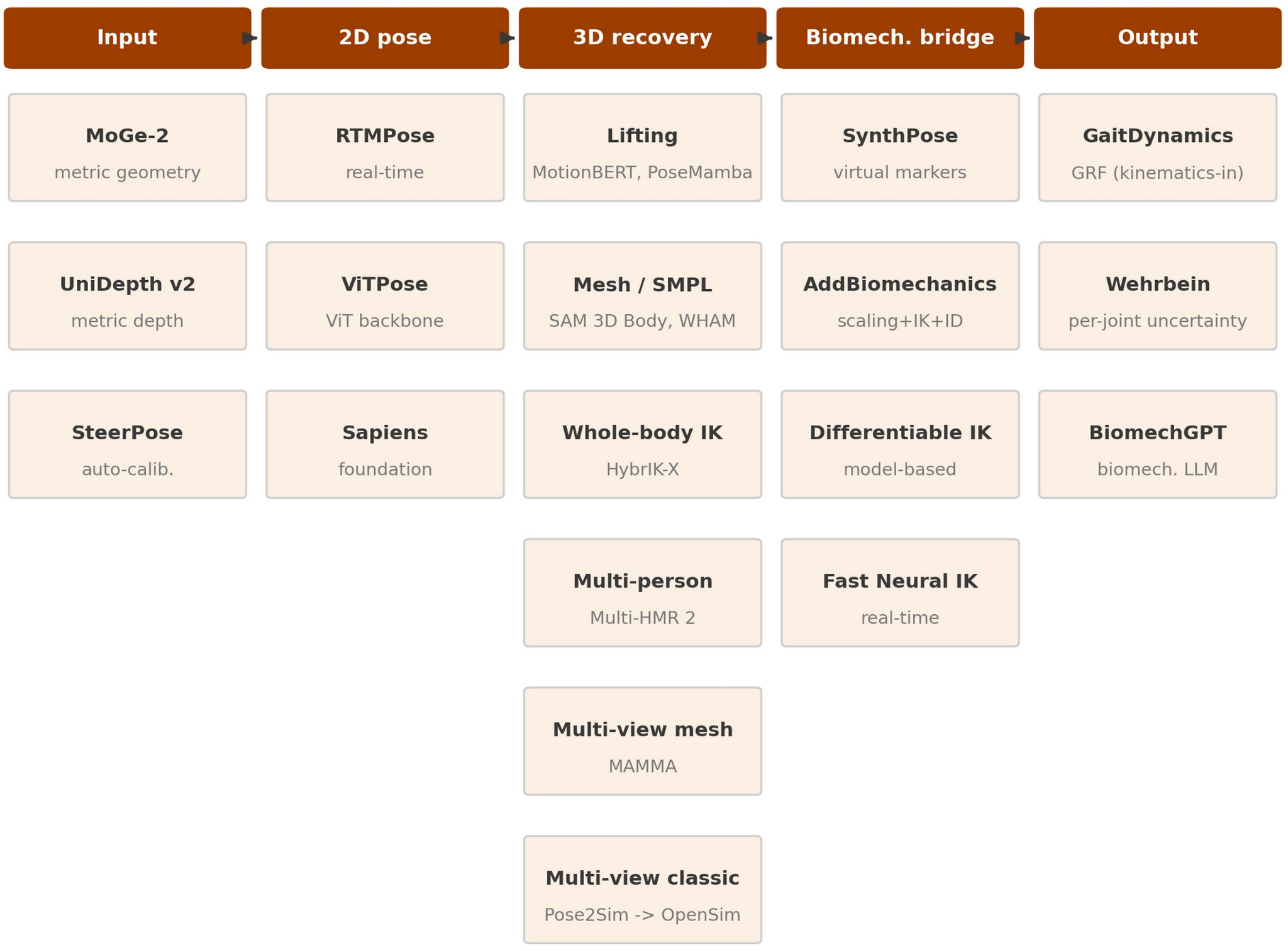

***Figure 7. The state-of-the-art toolbox: a candidate upgrade for each pipeline stage (one representative selection per stage, not exhaustive). Integrated pipelines (BioPose, OpenCap Monocular, Portable Biomechanics Laboratory) compose several of these blocks and are discussed in the text.***

### 3.5 The MACRO-MICRO bridge

Mapping the five validated families onto the emerging blocks (Fig. 8) charts the translational frontier family by family. Only multi-view triangulation is actively importing emerging blocks: through the OpenCap lineage it has taken up virtual-marker augmentation and inverse kinematics with musculoskeletal modelling, and a handful of studies reach toward video kinetics. The four monocular families have adopted almost nothing beyond 2D pose estimation and direct geometric angle computation. Across all five families, mesh recovery, differentiable inverse kinematics, video kinetics and human-centric foundation models are essentially absent. The map therefore has a near-empty right-hand side: the blocks that would most change clinical capability are the least adopted.

Four blocks define the gap. First, foundation-model mesh recovery, which could give monocular pipelines a dense, anatomically addressable representation, is absent from validated studies. Second, differentiable and neural inverse kinematics, which would replace fragile post-hoc IK, exists only in MICRO. Third, kinetics from video, the least-served MACRO output (Section 3.3.5), is an active MICRO block. Fourth, human-centric foundation models are not yet used in validated work. These blocks are the components of a single mesh-centred pipeline and define the design space for the framework proposed in Section 5.

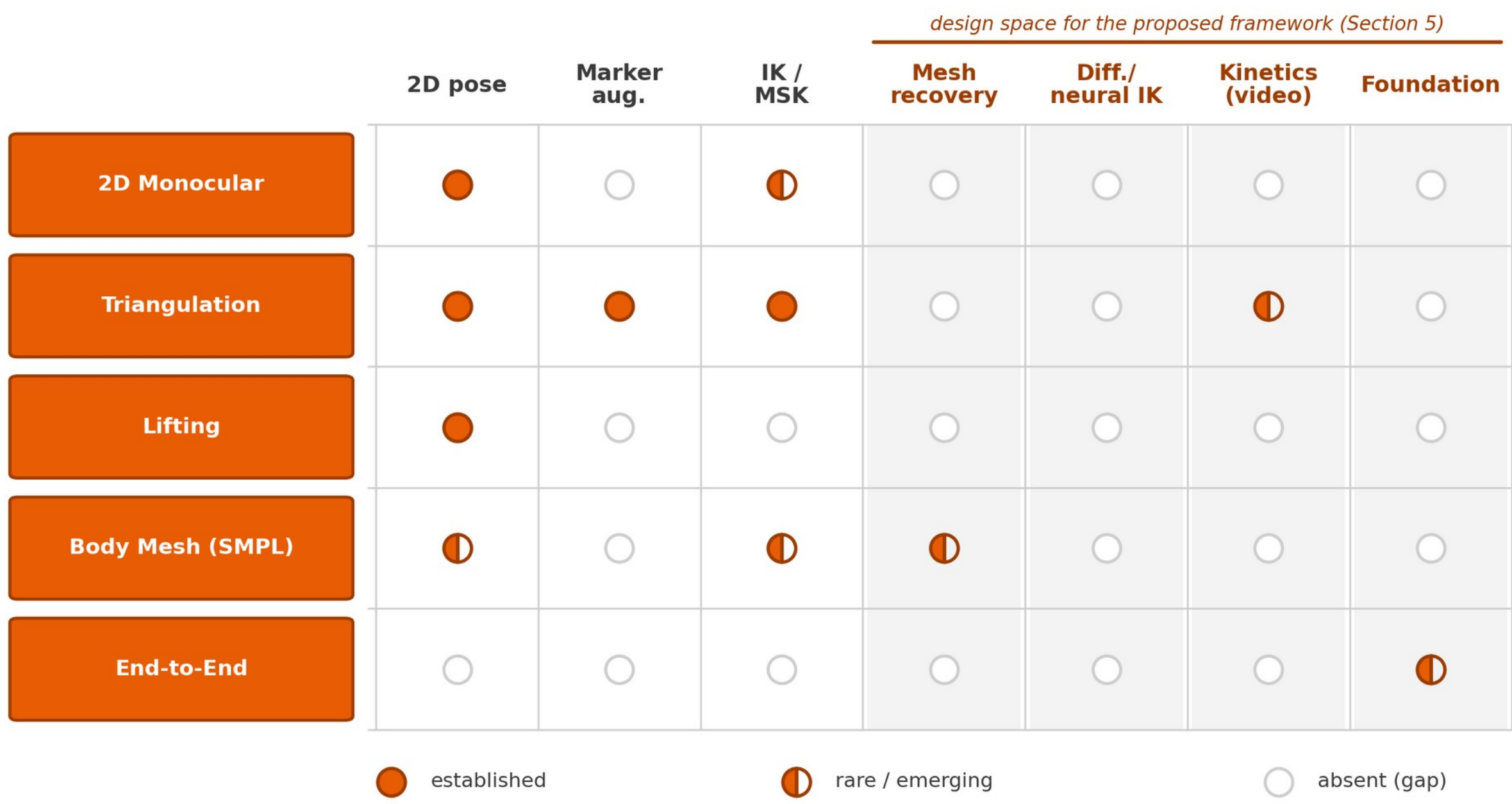

*Figure 8. Mapping between validated architectures (MACRO) and emerging building blocks (MICRO).*

# 4. Discussion

## 4.1 Summary of principal findings

This scoping review compared the validated literature on video-based markerless motion capture for biomechanics (the MACRO corpus) with the emerging computer-vision state of the art (the curated MICRO corpus). Five findings stand out. The field is young and growing fast, with most included studies published in the last two years. Its evidence base is narrow: participants are mostly healthy adults studied during gait, in the laboratory, with small samples and little real-time or out-of-laboratory use. The methods resolve into five architectural families across two modalities, monocular and multi-camera, yet most pipelines apply no biomechanical refinement and report raw geometric angles. Metrological performance falls short of the field's claim of marker-based equivalence: sagittal lower-limb agreement generally exceeds the error level McGinley et al. associate with likely clinical misinterpretation, while out-of-plane kinematics and kinetics are rarely validated. The building blocks most likely to change this picture, namely foundation-model mesh recovery, differentiable inverse kinematics and video kinetics, are active in computer vision but almost absent from validated biomechanics, consistent with the expected delay between computer-vision innovation and its validated biomedical uptake.

## 4.2 Maturity by modality and clinical translation

The two modalities trade deployability against metrological completeness. Multi-view triangulation is the more mature route to full three-dimensional biomechanics. It is the only family that routinely reports frontal and transverse kinematics, and it concentrates the inverse-kinematics and musculoskeletal pipelines, notably the OpenCap lineage, that make joint angles interpretable. It remains laboratory-bound and needs several synchronised, calibrated cameras. The monocular modality uses a single camera with no calibration and is far more deployable at the point of care. Yet most two-dimensional monocular studies stop at raw geometric angles (Section 3.2), and monocular lifting shows the largest errors (median lower-limb error around 6 to 7°). The clinical-acceptability gap therefore belongs to the field as a whole, not to one weak architecture. Spatiotemporal parameters such as walking speed agree closely with reference systems, but the point is narrower: claims of interchangeability with marker-based joint kinematics are premature for most pipelines.

## 4.3 Kinetics, the next frontier

The least-served output of the validated literature is kinetics. Ground-reaction forces and joint moments are tied most directly to joint loading, balance and fall risk, yet they were validated against force plates in only nine studies. They are arguably more clinically actionable than kinematics alone. An emerging MICRO block targets exactly this gap: physics-informed and learned estimation of forces and moments from video; generative foundation models now estimate ground-reaction forces from kinematics at near-laboratory accuracy [163]. The first single-camera-to-musculoskeletal-dynamics pipelines have appeared [12]. Force-plate-free kinetics is therefore an early and still-unproven prospect. It offers higher potential clinical value than incremental gains in sagittal joint-angle accuracy, but at a substantially higher validation cost, because it adds inverse-dynamics modelling error on top of the kinematic error and is hardest to validate beyond steady-state gait.

### 4.4 Gaps in the validated literature

Three gaps define the translational frontier. The first is plane coverage. Out-of-plane kinematics are scarcely reported (for the knee, 65 sagittal vs 5 frontal vs 3 transverse), so the measurements most relevant to many clinical questions are the least validated, and where reported they are larger. The second is populations. Clinical and pathological cohorts are rare, and older-adult and paediatric cohorts rarer still, although these groups would benefit most from accessible movement analysis. The third is operating conditions. Real-time analysis (7% of studies) and home or field capture (almost absent) remain largely unaddressed, although both are prerequisites for analysis outside the laboratory. Closing these gaps is a higher priority than further benchmarking of healthy-adult sagittal gait.

### 4.5 Strengths and limitations of this review

The review has two main strengths. It combines a registered PRISMA-ScR protocol with a broad three-database search and a dual-tier design that places the validated literature within the trajectory of the underlying computer-vision research. Several limitations apply. By design the review excludes black-box commercial systems (Section 2.3), including Theia3D, among the most clinically deployed, and RGB-D depth sensors that remain common in rehabilitation; the findings therefore characterise the validated open-architecture evidence base, not the full deployed clinical landscape. The MICRO corpus is curated and non-exhaustive, so its block counts reflect screening and publication volume, not biomechanical importance, and we used them only qualitatively. Data extraction used a large-language-model-assisted pipeline with cross-vendor checking and adjudication, validated by manual re-extraction of a random 20% of the included studies, and we report only cleaned values, excluding signed-bias, velocity-derived and wrong-metric entries. As a scoping review, the synthesis is structured and narrative, not meta-analytic, and we did not formally appraise risk of bias. We benchmarked against McGinley et al.'s error bands, 2° or less acceptable, 2 to 5° reasonable, above 5° likely to mislead [131]; acceptable error ultimately depends on the application, so this band analysis is a reference point, not an absolute cut-off.

## 4.6 Implications for clinical practice and rehabilitation research

The usable output today depends on the measurement (Table 3). Spatiotemporal parameters such as walking speed, step length and step time agree closely with reference systems and are appropriate for monitoring, screening and remote assessment; walking speed in particular is described as a sixth vital sign in older adults [165]. Sagittal joint angles are usable for gross qualitative trends but not for decisions that need resolution below 3 to 5°, such as tracking a small rehabilitation change. Frontal and transverse angles and all kinetics remain research-grade and are not yet validated for clinical use.

**Table 3. Clinical readiness of markerless motion-capture outputs.**

| Readiness tier | Outputs | Appropriate use |
|---|---|---|
| Ready for clinical use | Spatiotemporal parameters (walking speed, step length, step time) | Monitoring, screening and remote assessment |
| Ready for screening and trends | Sagittal-plane joint angles (gross flexion-extension) | Qualitative trend tracking; not decisions needing resolution below three to five degrees |
| Research-grade only | Frontal- and transverse-plane joint angles; force-plate-free kinetics; pathological-gait profiling | Not to be used for clinical decisions |

A safety point follows from the structure of the field. The most deployable tools, single-camera monocular pipelines, are also the least refined, since 46 of 47 report raw geometric angles. Deployability and validation-readiness are therefore inversely related: a cheap monocular tool may reach clinics precisely because it is accessible, before it has been validated for the decision it is used for.

Healthy-adult validation does not transfer to patients. Pathological gait violates the assumptions these pipelines are built on: assistive devices such as walkers, orthoses and canes; atypical morphology; involuntary movement such as tremor, spasticity and dyskinesia; and severe out-of-plane deviation such as circumduction and scissoring. A 5° healthy-adult error is not a safe prior for a hemiparetic or crouch-gait patient. Population-specific validation, not simply more participants, is required. Each of these breaks a specific assumption. Assistive devices and orthoses occlude or displace the landmarks pose estimators depend on. Involuntary movement violates the temporal-smoothness priors used to denoise trajectories. The largest pathological deviations are frontal and transverse, the planes these systems estimate worst, so the clinical signal and the measurement error fall in the same plane.

Three practical risks deserve attention. Pose-estimation training and validation data skew toward young, lean, light-skinned adults, the same skew this review documents, so performance may be worse in the populations least represented, which is a fairness concern. Clinical and home video raises privacy and consent questions absent in a marker laboratory. And governance must keep validation ahead of deployment.

## 5. Future design guidelines for clinically deployable markerless biomechanics

This section delivers the third objective: translating the validated MACRO lessons and the emerging MICRO blocks into design guidelines for a clinically deployable pipeline. These are hypothesis-generating directions, not a validated method, each tied to a gap documented in Sections 3 and 4 rather than restated here. One principle governs them all. For the clinical target populations, older and pathological cohorts, generalisability must come from what is universal, anatomy, mechanics and a subject-specific model, not from priors learned on populations that exclude them. The direction therefore combines established components (subject-specific OpenSim scaling and inverse kinematics, and physics-based refinement) with emerging ones (dense mesh recovery; differentiable, model-based inverse kinematics; per-degree-of-freedom uncertainty) and some untested in validated biomechanics (multi-view mesh fusion and a two-level mesh-to-anatomy correspondence). Any learned prior is kept at a stage where its errors are checkable against the image and the component is replaceable, so that final fidelity rests on subject-specific anatomy and mechanics. All components are drawn from the published literature and are openly available with public code. The perspective proposes no proprietary system. We distil five guidelines (Fig. 9).

First, we propose a dense parametric mesh front-end, recovering a dense parametric human mesh per frame rather than sparse keypoints and using human-mesh foundation models absent from the validated corpus (for example SAM 3D Body [155] and its real-time variant [166], building on the Sapiens line [153], with clinical-landmark transfer shown by MedSapiens [167]). A dense mesh carries segment orientation and out-of-plane rotation, precisely what the validated monocular families rarely report. Its estimator inherits an able-bodied population bias, but this risk is bounded: the estimator is an interchangeable component that improves as models diversify, and a mesh error is local and checkable against the image, unlike a silent regression inside a learned kinematic model. This addresses the first gap of Section 3.5: mesh recovery is present in MICRO but absent from every validated MACRO study.

Second, we address the step from mesh to clinically defined anatomy. A parametric mesh has no clinically defined joint centres, and reading virtual markers from fixed mesh positions conflates a generic, calibratable mesh-to-anatomy offset with small individual variation. Because imprecise registration of one to two centimetres can induce five to ten degrees of joint-angle error, this matters most for atypical populations, whose clinically meaningful deviations are themselves small. We propose a two-level correspondence: a generic mesh-to-landmark map calibrated once on an annotated reference and then frozen, plus a small, strongly regularised subject-specific residual, bony landmarks constrained more tightly than soft-tissue ones. Neural Localizer Fields [158] make this feasible by allowing any anatomical point to be queried on the mesh. This addresses a gap MICRO leaves open: virtual markers are read from fixed positions, without the calibration atypical populations require.

Third, kinematic inference should be subject-specific, model-based and not learned. This is arguably the most consequential choice for clinical translation. Three options exist. A learned neural inverse kinematics predicts angles from a population-trained network (BioPose [11], Fast Neural IK [162]); its generalisability is bounded by the training distribution. A differentiable inverse kinematics is the same model-based solve rendered differentiable, so scaling, registration and inverse kinematics are optimised jointly by gradient descent (Differentiable Biomechanics [160], JAX-IK [161], and AddBiomechanics [168], which also solves inverse dynamics); differentiable does not mean learned. A classical model-based solve (OpenSim) fits joint angles subject by subject under anatomical constraints, with no population training. For older and pathological populations the distinction may be decisive: an inverse kinematics learned on able-bodied movement may regress toward its training patterns and mask the pathological deviations the measurement should capture. We therefore suggest kinematic inference rest on subject-specific, model-based inverse kinematics, optionally made differentiable, rather than on a learned mapping. This addresses the second gap of Section 3.5, where differentiable and neural inverse kinematics appear only in MICRO, and whether a learned solve transfers to the target populations remains open.

Fourth, we distinguish two deployment modalities. In the monocular case (one camera, the most deployable), per-frame meshes are consolidated into one morphology per trial through a temporal shape-lock that fixes segment lengths; because soft-tissue artefact also varies marker-based segment lengths, a shape-locked estimate could on this axis approach the consistency of marker-based capture, a testable hypothesis [117]. A physics-based refinement stage (foot contact, anti-drift, smoothing) then counters monocular drift, in the lineage of OpenCap Monocular [12]. In the multi-camera case, rather than triangulating sparse keypoints, a single mesh is fitted jointly to all calibrated views by visibility-weighted reprojection; the fixed topology makes inter-view correspondence intrinsic and lets several views observe segment lengths directly. This mesh-fusion route, inspired by MAMMA [169], is essentially unexplored in validated biomechanics, with hypothesised gains (occlusion robustness, out-of-plane recovery) still to be tested (gaps of Sections 3.3.4 and 4.2).

Fifth, we turn to outputs, kinetics and uncertainty. Both modalities target joint kinematics across all degrees of freedom. Kinetics without force plates is the field's least-validated output (Section 3.3.5) and remains a distant prospect: generative foundation models already estimate ground-reaction forces from kinematics [163], but their generalisation to atypical clinical cohorts is untested and, like learned inverse kinematics, they would need training data covering older and pathological populations to be reliable there. Per-degree-of-freedom uncertainty, almost absent from the validated literature, is a more immediate and clinically essential addition, since whether a measurement is reliable conditions the clinical decision; probabilistic pose estimation with normalizing flows (Wehrbein et al. [164]) offers one route. Both are among the least-validated outputs identified in Section 3.3.5.

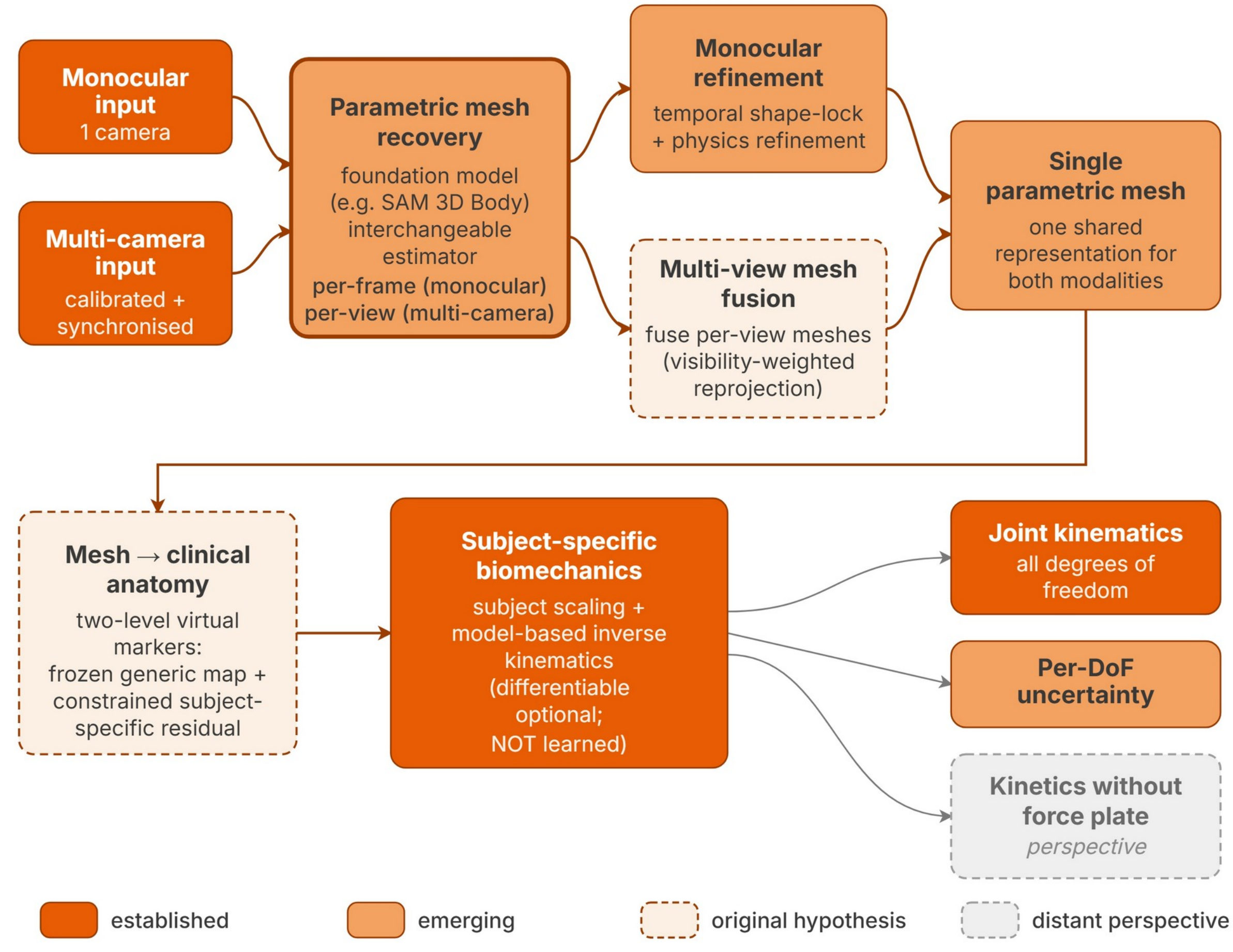

*Figure 9. Research agenda for clinically deployable markerless biomechanics, shown as a candidate pipeline. Monocular and multi-camera paths share one parametric-mesh engine and rejoin into a single mesh, mapped to clinical anatomy through virtual markers and solved by subject-specific inverse kinematics. Outputs are joint kinematics with per-degree-of-freedom uncertainty, plus force-plate kinetics in the multi-camera setting.*

## 6. Conclusions

Video-based markerless motion capture has matured quickly, but the validated evidence does not yet support interchangeability with marker-based systems for clinical joint kinematics. The capabilities most relevant to clinical biomechanics, namely out-of-plane kinematics, kinetics, and validation in pathological and older populations under real-world conditions, are the least developed, while the computer-vision advances that could address them have not entered the validated literature. We distil these documented limitations into evidence-based design guidelines, each mapping a gap onto a corresponding emerging block, offered as perspectives to be tested rather than a validated method. Progress will depend on standardised, biomechanically grounded evaluation that reports joint angles rather than keypoint-position error, on validation in the populations and settings that matter clinically, and on closing the lag between computer-vision innovation and biomedical validation.

## Declarations

**Ethics approval and consent to participate.** Not applicable (secondary analysis of published literature).

**Consent for publication.** Not applicable.

**Availability of data and materials.** The protocol and the complete extraction dataset are available in the Open Science Framework repository (DOI 10.17605/OSF.IO/TBYPC); the extraction table is provided as a machine-readable Additional file.

**Competing interests.** The authors declare no competing interests.

**Funding.** FD is supported by a doctoral fellowship from the Doctoral School ED SMH (Sciences du Mouvement Humain), Université Côte d'Azur. This work was supported in part by the PRESAGE project (French National Research Agency, ANR-23-PAVH-0002), which funded review resources including the Covidence systematic-review platform. The funders had no role in the design, conduct, analysis, or reporting of the review.

**Authors' contributions.** FD: conceptualization, methodology, software, formal analysis, investigation, data curation, visualization, writing – original draft. EP: investigation (independent, blinded study screening and selection on Covidence), writing – review and editing. FC: writing – review and editing. RZ: validation (adjudication of screening conflicts), supervision, funding acquisition, writing – review and editing. All authors read and approved the final manuscript.

**Acknowledgements.** Not applicable.

## References

1. Windolf M, Götzen N, Morlock M. Systematic accuracy and precision analysis of video motion capturing systems: exemplified on the Vicon-460 system. J Biomech. 2008;41(12):2776-2780. https://doi.org/10.1016/j.jbiomech.2008.06.024

2. Leardini A, Chiari L, Della Croce U, Cappozzo A. Human movement analysis using stereophotogrammetry. Part 3: soft tissue artifact assessment and compensation. Gait Posture. 2005;21(2):212-225. https://doi.org/10.1016/j.gaitpost.2004.05.002

3. Baker R. Gait analysis methods in rehabilitation. J Neuroeng Rehabil. 2006;3(1):4. https://doi.org/10.1186/1743-0003-3-4

4. Colyer SL, Evans M, Cosker DP, Salo AIT. A Review of the Evolution of Vision-Based Motion Analysis and the Integration of Advanced Computer Vision Methods Towards Developing a Markerless System. Sports Med Open. 2018;4(1):24. https://doi.org/10.1186/s40798-018-0139-y

5. Cao Z, Hidalgo G, Simon T, Wei SE, Sheikh Y. OpenPose: realtime multi-person 2D pose estimation using part affinity fields. IEEE Trans Pattern Anal Mach Intell. 2021;43(1):172-186. https://doi.org/10.1109/TPAMI.2019.2929257

6. Bazarevsky V, Grishchenko I, Raveendran K, Zhu T, Zhang F, Grundmann M. BlazePose: on-device real-time body pose tracking. arXiv:2006.10204. 2020.

7. Sun K, Xiao B, Liu D, Wang J. Deep high-resolution representation learning for human pose estimation. In: Proc IEEE/CVF Conf Comput Vis Pattern Recognit (CVPR). 2019. p. 5686-5696. https://doi.org/10.1109/CVPR.2019.00584

8. Pagnon D, Domalain M, Reveret L. Pose2Sim: An End-to-End Workflow for 3D Markerless Sports Kinematics-Part 1: Robustness. Sensors. 2021;21(19):6530. https://doi.org/10.3390/s21196530

9. Uhlrich SD, Falisse A, Kidzinski L, Muccini J, Ko M, Chaudhari AS, et al. OpenCap: Human movement dynamics from smartphone videos. PLOS Computational Biology. 2023;19(10):e1011462. https://doi.org/10.1371/journal.pcbi.1011462

10. Zhu W, Ma X, Liu Z, Liu L, Wu W, Wang Y. MotionBERT: A Unified Perspective on Learning Human Motion Representations. In: 2023 IEEE/CVF International Conference on Computer Vision (ICCV). 2023. p. 15039-15053. https://doi.org/10.1109/iccv51070.2023.01385

11. Koleini F, Saleem MU, Wang P, Xue H, Helmy A, Fenwick A. BioPose: Biomechanically-Accurate 3D Pose Estimation from Monocular Videos. In: 2025 IEEE/CVF Winter Conference on Applications of Computer Vision (WACV). IEEE; 2025. arXiv:2501.07800 [cs.CV].

12. Gilon S, Miller EY, Uhlrich SD. OpenCap Monocular: 3D Human Kinematics and Musculoskeletal Dynamics from a Single Smartphone Video. arXiv:2603.24733 [cs.CV]. 2026.

13. Balaban B, Tok F. Gait disturbances in patients with stroke. PM R. 2014;6(7):635-642. https://doi.org/10.1016/j.pmrj.2013.12.017

14. Lam WWT, Tang YM, Fong KNK. A systematic review of the applications of markerless motion capture (MMC) technology for clinical measurement in rehabilitation. J Neuroeng Rehabil. 2023;20(1):57. https://doi.org/10.1186/s12984-023-01186-9

15. Tricco AC, Lillie E, Zarin W, O'Brien KK, Colquhoun H, Levac D, et al. PRISMA Extension for Scoping Reviews (PRISMA-ScR): Checklist and Explanation. Ann Intern Med. 2018;169(7):467-473. https://doi.org/10.7326/M18-0850

16. Peters MDJ, Marnie C, Tricco AC, Pollock D, Munn Z, Alexander L, et al. Updated methodological guidance for the conduct of scoping reviews. JBI Evid Synth. 2020;18(10):2119-2126. https://doi.org/10.11124/jbies-20-00167

17. Young F, Mason R, Morris R, Stuart S, Godfrey A. Internet-of-Things-Enabled Markerless Running Gait Assessment from a Single Smartphone Camera. Sensors. 2023;23(2):696. https://doi.org/10.3390/s23020696

18. Verhoeven M, Zandvoort CS, Dominici N. From marker to markerless: Validating DeepLabCut for 2D sagittal plane gait analysis in adults and newly walking toddlers. Journal of Biomechanics. 2025;186:112708. https://doi.org/10.1016/j.jbiomech.2025.112708

19. Takeda I, Yamada A, Onodera H. Artificial Intelligence-Assisted motion capture for medical applications: a comparative study between markerless and passive marker motion capture. Computer Methods in Biomechanics and Biomedical Engineering. 2020;24(8):864-873. https://doi.org/10.1080/10255842.2020.1856372

20. Stenum J, Rossi C, Roemmich RT. Two-dimensional video-based analysis of human gait using pose estimation. PLOS Computational Biology. 2021;17(4):e1008935. https://doi.org/10.1371/journal.pcbi.1008935

21. Molteni LE, Andreoni G. Comparing the Accuracy of Markerless Motion Analysis and Optoelectronic System for Measuring Gait Kinematics of Lower Limb. Bioengineering. 2025;12(4):424. https://doi.org/10.3390/bioengineering12040424

22. Sabo A, Gorodetsky C, Fasano A, Iaboni A, Taati B. Concurrent validity of Zeno instrumented walkway and video-based gait features in adults with Parkinson's disease. IEEE Journal of Translational Engineering in Health and Medicine. 2022;10:1-11. https://doi.org/10.1109/jtehm.2022.3180231

23. Panconi G, Grasso S, Guarducci S, Mucchi L, Minciacchi D, Bravi R. DeepLabCut custom-trained model and the refinement function for gait analysis. Scientific Reports. 2025;15(1):2364. https://doi.org/10.1038/s41598-025-85591-1

24. Ota M, Tateuchi H, Hashiguchi T, Ichihashi N. Verification of validity of gait analysis systems during treadmill walking and running using human pose tracking algorithm. Gait & Posture. 2021;85:290-297. https://doi.org/10.1016/j.gaitpost.2021.02.006

25. Moro M, Marchesi G, Odone F, Casadio M. Markerless gait analysis in stroke survivors based on computer vision and deep learning: a pilot study. Proceedings of the 35th Annual ACM Symposium on Applied Computing. 2020:2097-2104. https://doi.org/10.1145/3341105.3373963

26. Menychtas D, Petrou N, Kansizoglou I, Giannakou E, Grekidis A, Gasteratos A, et al. Gait analysis comparison between manual marking, 2D pose estimation algorithms, and 3D marker-based system. Frontiers in Rehabilitation Sciences. 2023;4:1238134. https://doi.org/10.3389/fresc.2023.1238134

27. Mehdizadeh S, Nabavi H, Sabo A, Arora T, Iaboni A, Taati B. Concurrent validity of human pose tracking in video for measuring gait parameters in older adults: a preliminary analysis with multiple trackers, viewing angles, and walking directions. Journal of NeuroEngineering and Rehabilitation. 2021;18(1):139. https://doi.org/10.1186/s12984-021-00933-0

28. Lonini L, Moon Y, Embry K, Cotton RJ, McKenzie K, Jenz S, et al. Video-Based Pose Estimation for Gait Analysis in Stroke Survivors during Clinical Assessments: A Proof-of-Concept Study. Digital Biomarkers. 2022;6(1):9-18. https://doi.org/10.1159/000520732

29. Ino T, Samukawa M, Ishida T, Wada N, Koshino Y, Kasahara S, et al. Validity of AI-Based Gait Analysis for Simultaneous Measurement of Bilateral Lower Limb Kinematics Using a Single Video Camera. Sensors. 2023;23(24):9799. https://doi.org/10.3390/s23249799

30. Lambricht N, Englebert A, Nguyen AP, Fisette P, Pitance L, Detrembleur C. Impact of Running Clothes on Accuracy of Smartphone-Based 2D Joint Kinematic Assessment During Treadmill Running Using OpenPifPaf. Sensors. 2025;25(3):934. https://doi.org/10.3390/s25030934

31. Lambricht N, Englebert A, Pitance L, Fisette P, Detrembleur C. Quantifying performance and joint kinematics in functional tasks crucial for anterior cruciate ligament rehabilitation using smartphone video and pose detection. The Knee. 2025;52:171-178. https://doi.org/10.1016/j.knee.2024.11.006

32. John K, Stenum J, Chiang CC, French MA, Kim C, Manor J, et al. Accuracy of Video-Based Gait Analysis Using Pose Estimation During Treadmill Walking Versus Overground Walking in Persons After Stroke. Physical Therapy. 2023;104(2):pzad121. https://doi.org/10.1093/ptj/pzad121

33. Ino T, Samukawa M, Ishida T, Wada N, Koshino Y, Kasahara S, et al. Validity and Reliability of OpenPose-Based Motion Analysis in Measuring Knee Valgus during Drop Vertical Jump Test. Journal of Sports Science and Medicine. 2024;:515-525. https://doi.org/10.52082/jssm.2024.515

34. Dunn M, Kennerley A, Murrell-Smith Z, Webster K, Middleton K, Wheat J. Application of video frame interpolation to markerless, single-camera gait analysis. Sports Engineering. 2023;26(1):22. https://doi.org/10.1007/s12283-023-00419-3

35. Drazan JF, Phillips WT, Seethapathi N, Hullfish TJ, Baxter JR. Moving outside the lab: Markerless motion capture accurately quantifies sagittal plane kinematics during the vertical jump. Journal of Biomechanics. 2021;125:110547. https://doi.org/10.1016/j.jbiomech.2021.110547

36. Boldo M, Di Marco R, Martini E, Nardon M, Bertucco M, Bombieri N. On the reliability of single-camera markerless systems for overground gait monitoring. Computers in Biology and Medicine. 2024;171:108101. https://doi.org/10.1016/j.compbiomed.2024.108101

37. Andreoni G, Molteni LE. Comparison of the Accuracy of Markerless Motion Analysis and Optoelectronic System for Measuring Lower Limb Gait Kinematics. Lecture Notes in Computer Science. 2024;:3-15. https://doi.org/10.1007/978-3-031-61063-9_1

38. Anderson JT, Stenum J, Roemmich RT, Wilson RB. Validation of markerless video-based gait analysis using pose estimation in toddlers with and without neurodevelopmental disorders. Frontiers in Digital Health. 2025;7. https://doi.org/10.3389/fdgth.2025.1542012

39. Aleksic J, Kanevsky D, Mesaroš D, Knezevic OM, Cabarkapa D, Bozovic B, et al. Validation of Automated Countermovement Vertical Jump Analysis: Markerless Pose Estimation vs. 3D Marker-Based Motion Capture System. Sensors. 2024;24(20):6624. https://doi.org/10.3390/s24206624

40. Yagi K, Sugiura Y, Hasegawa K, Saito H. Gait Measurement at Home Using A Single RGB Camera. Gait & Posture. 2020;76:136-140. https://doi.org/10.1016/j.gaitpost.2019.10.006

41. Tahara AK, Chinaglia AG, Monteiro RLM, Bedo BLS, Cesar GM, Santiago PRP. Predicting walkway spatiotemporal parameters using a markerless, pixel-based machine learning approach. Brazilian Journal of Motor Behavior. 2025;19(1):e462. https://doi.org/10.20338/bjmb.v19i1.462

42. Francia C, Donno L, Chiosso S, Palotti G, Tarabini M, Galli M. Dual-Camera System for AI Markerless 3D Lower Limb Motion Analysis. Lecture Notes in Computer Science. 2025;:303-312. https://doi.org/10.1007/978-3-031-97781-7_22

43. Yamamoto M, Shimatani K, Hasegawa M, Kurita Y, Ishige Y, Takemura H. Accuracy of Temporo-Spatial and Lower Limb Joint Kinematics Parameters Using OpenPose for Various Gait Patterns With Orthosis. IEEE Transactions on Neural Systems and Rehabilitation Engineering. 2021;29:2666-2675. https://doi.org/10.1109/tnsre.2021.3135879

44. Wade L, Needham L, Evans M, McGuigan P, Colyer S, Cosker D, et al. Examination of 2D frontal and sagittal markerless motion capture: Implications for markerless applications. PLOS ONE. 2023;18(11):e0293917. https://doi.org/10.1371/journal.pone.0293917

45. Kim J, Kim R, Byun K, Kang N, Park K. Assessment of temporospatial and kinematic gait parameters using human pose estimation in patients with Parkinson's disease: A comparison between near-frontal and lateral views. PLOS ONE. 2025;20(1):e0317933. https://doi.org/10.1371/journal.pone.0317933

46. Balci IC, Sayin I, Salturk S, Gursoy R, Ozsoy U, Dogru HC, et al. Reliability assessment of markerless technologies in biomechanical motion analysis: a performance comparison. Frontiers in Sports and Active Living. 2026;7. https://doi.org/10.3389/fspor.2025.1712332

47. Hii CST, Gan KB, Zainal N, Mohamed Ibrahim N, Azmin S, Mat Desa SH, et al. Automated Gait Analysis Based on a Marker-Free Pose Estimation Model. Sensors. 2023;23(14):6489. https://doi.org/10.3390/s23146489

48. Ge F, Wu C, Ge F, Xu S, Xiao J. Reliability and validity of OpenPose for measuring HKA angle in dynamic walking videos in patients with knee osteoarthritis. Scientific Reports. 2025;15(1). https://doi.org/10.1038/s41598-025-09627-2

49. Asaeda M, Onishi T, Ito H, Miyahara S, Mikami Y. Reliability and validity of knee valgus angle calculation at single-leg drop landing by posture estimation using machine learning. Heliyon. 2024;10(17):e36338. https://doi.org/10.1016/j.heliyon.2024.e36338

50. Haberkamp LD, Garcia MC, Bazett-Jones DM. Validity of an artificial intelligence, human pose estimation model for measuring single-leg squat kinematics. Journal of Biomechanics. 2022;144:111333. https://doi.org/10.1016/j.jbiomech.2022.111333

51. Washabaugh EP, Shanmugam TA, Ranganathan R, Krishnan C. Comparing the accuracy of open-source pose estimation methods for measuring gait kinematics. Gait & Posture. 2022;97:188-195. https://doi.org/10.1016/j.gaitpost.2022.08.008

52. Gao X, Cheng X, Jiao Y, Reading S, Zhang Y. Can Gait Deviations Be Identified through Video-Based Gait Analysis? Validation of the BlazePose Pose Estimation Algorithm. 2025 47th Annual International Conference of the IEEE Engineering in Medicine and Biology Society (EMBC). 2025:1-5. https://doi.org/10.1109/embc58623.2025.11254531

53. Kumthekar PS, Sharma A, Malla S, Richardson RT, Morales AW, Nguyen H, et al. A Pilot Study for Developing Mobile App and Cloud Computing for Upper Extremities Motion Analysis. 2024 IEEE Cloud Summit. 2024:24-27. https://doi.org/10.1109/cloud-summit61220.2024.00011

54. Nishizawa K, Oba Y, Yamada K, Tanaka I, Tsumugiwa T, Yokogawa R, Watanabe T. Evaluation of the clinical utility of a gait analysis system using pose estimation techniques in physical therapy. In: 2024 SICE International Symposium on Control Systems (SICE ISCS). Higashi-Hiroshima, Japan; 2024. p. 107-10.

55. Hii CST, Gan KB, You HW, Zainal N, Ibrahim NM, Azmin S, et al. Frontal Plane Gait Analysis using Pose Estimation Models. 2023 IEEE 2nd National Biomedical Engineering Conference (NBEC). 2023:1-6. https://doi.org/10.1109/nbec58134.2023.10352623

56. Irfan M, Suryadevara NK. Reliable Gait Measurements Using Smartphone Vision Sensor. IEEE Sensors Journal. 2025;25(1):1487-1494. https://doi.org/10.1109/jsen.2024.3487541

57. Ceriola L, Taborri J, Donati M, Rossi S, Patanè F, Mileti I. Comparative Analysis of Markerless Motion Capture Systems for Measuring Human Kinematics. IEEE Sensors Journal. 2024;24(17):28135-28144. https://doi.org/10.1109/jsen.2024.3431873

58. Kidziński Ł, Yang B, Hicks JL, Rajagopal A, Delp SL, Schwartz MH. Deep neural networks enable quantitative movement analysis using single-camera videos. Nature Communications. 2020;11(1). https://doi.org/10.1038/s41467-020-17807-z

59. Hwang W, Shim D, Kim J, Oh KR, Chung SG, Beom J, et al. Sit-to-Stand Power From 2D Pose Estimation as an Indicator of Muscle Strength in Older Adults. Journal of Cachexia, Sarcopenia and Muscle. 2026;17(1):e70208. https://doi.org/10.1002/jcsm.70208

60. Irfan M, Suryadevara NK, Biswas R, Gaddam A. Kinematic Gait Analysis Using Markerless System to Determine Joint Angles. Lecture Notes in Networks and Systems. 2024;:551-559. https://doi.org/10.1007/978-981-97-2671-4_42

61. Talaa S, Jilbab A, El Yousfi Alaoui MH, Ihssane K. Markerless Knee Angle Measurement: Advancements in Gait Analysis Using Computer Vision. Lecture Notes in Networks and Systems. 2025;:103-112. https://doi.org/10.1007/978-3-031-94623-3_9

62. Ang YH, Chan CK, Yap SC, Toa CK, Tran P, Goh SK. Markerless Human Motion Analysis for Telerehabilitation: A Case Study on Squat. Advances in Science, Technology & Innovation. 2024;:249-259. https://doi.org/10.1007/978-3-031-52303-8_18

63. Tsintzira K, Smyrli A, Mastrogeorgiou A, Papadopoulos E. Low-Cost Markerless Gait Analysis Using a Minimal Human Model. 2025 IEEE 25th International Conference on Bioinformatics and Bioengineering (BIBE). 2025:558-562. https://doi.org/10.1109/BIBE66822.2025.00099

64. Natraj S, Messmer T, Fujii Y, Suzuki K, Riener R, Eriks-Hoogland I, et al. 3D pose estimation for scalable remote gait kinematics assessment. npj Digital Medicine. 2025;9(1):37. https://doi.org/10.1038/s41746-025-02211-y

65. Zhu X, Boukhennoufa I, Liew B, Gao C, Yu W, McDonald-Maier KD, et al. Monocular 3D Human Pose Markerless Systems for Gait Assessment. Bioengineering. 2023;10(6):653. https://doi.org/10.3390/bioengineering10060653

66. Mercadal-Baudart C, Liu CJ, Farrell G, Boyne M, Gonzalez Escribano J, Smolic A, et al. Exercise quantification from single camera view markerless 3D pose estimation. Heliyon. 2024;10(6):e27596. https://doi.org/10.1016/j.heliyon.2024.e27596

67. Lagomarsino B, Massone A, Odone F, Casadio M, Moro M. Video-based markerless assessment of bilateral upper limb motor activity following cervical spinal cord injury. Computers in Biology and Medicine. 2025;196:110908. https://doi.org/10.1016/j.compbiomed.2025.110908

68. Carrasco-Plaza J, Cerda M. Evaluation of Human Pose Estimation in 3D with Monocular Camera for Clinical Application. Communications in Computer and Information Science. 2022;:121-134. https://doi.org/10.1007/978-3-030-98457-1_10

69. Barzyk P, Zimmermann P, Stein M, Keim D, Gruber M. AI-smartphone markerless motion capturing of hip, knee, and ankle joint kinematics during countermovement jumps. European Journal of Sport Science. 2024;24(10):1452-1462. https://doi.org/10.1002/ejsc.12186

70. Barzyk P, Boden AS, Howaldt J, Stürner J, Zimmermann P, Seebacher D, et al. Steps to Facilitate the Use of Clinical Gait Analysis in Stroke Patients: The Validation of a Single 2D RGB Smartphone Video-Based System for Gait Analysis. Sensors. 2024;24(23):7819. https://doi.org/10.3390/s24237819

71. Azhand A, Rabe S, Müller S, Sattler I, Heimann-Steinert A. Algorithm based on one monocular video delivers highly valid and reliable gait parameters. Scientific Reports. 2021;11(1). https://doi.org/10.1038/s41598-021-93530-z

72. Wang H, Su B, Lu L, Jung S, Qing L, Xie Z, et al. Markerless gait analysis through a single camera and computer vision. Journal of Biomechanics. 2024;165:112027. https://doi.org/10.1016/j.jbiomech.2024.112027

73. Ueno R. Calibrationless monocular vision musculoskeletal simulation during gait. Heliyon. 2024;10(11):e32078. https://doi.org/10.1016/j.heliyon.2024.e32078

74. Hulleck AA, AlShehhi A, El Rich M, Khan R, Katmah R, Mohseni M, et al. BlazePose-Seq2Seq: Leveraging Regular RGB Cameras for Robust Gait Assessment. IEEE Transactions on Neural Systems and Rehabilitation Engineering. 2024;32:1715-1724. https://doi.org/10.1109/tnsre.2024.3391908

75. Wang XM, Smith DT, Zhu Q. A webcam-based machine learning approach for three-dimensional range of motion evaluation. PLOS ONE. 2023;18(10):e0293178. https://doi.org/10.1371/journal.pone.0293178

76. He R, You Z, Zhou Y, Chen G, Diao Y, Jiang X, et al. A novel multi-level 3D pose estimation framework for gait detection of Parkinson’s disease using monocular video. Frontiers in Bioengineering and Biotechnology. 2024;12. https://doi.org/10.3389/fbioe.2024.1520831

77. Liang S, Zhang Y, Diao Y, Li G, Zhao G. The reliability and validity of gait analysis system using 3D markerless pose estimation algorithms. Frontiers in Bioengineering and Biotechnology. 2022;10. https://doi.org/10.3389/fbioe.2022.857975

78. Lin ZY, Lyu B, Fernandez JC, van der Kruk E, Seth A, Zhang X. 3D Kinematics Estimation from Video with a Biomechanical Model and Synthetic Training Data. 2024 IEEE/CVF Conference on Computer Vision and Pattern Recognition Workshops (CVPRW). 2024;:1441-1450. https://doi.org/10.1109/cvprw63382.2024.00151

79. Pappas MC, Boughanem DJ, Baudendistel ST, Chen S, Liu S, Acevedo GT, et al. Video-based Clinical Gait Analysis in Parkinson's Disease: A Novel Approach Using Frontal Plane Videos and Machine Learning. 2024 46th Annual International Conference of the IEEE Engineering in Medicine and Biology Society (EMBC). 2024:1-4. https://doi.org/10.1109/embc53108.2024.10781504

80. Rode D, Dunkel A, Willi R, Wolf P, Xiloyannis M, Riener R. Assessment of monocular human pose estimation models for clinical movement analysis. Scientific Reports. 2025;15(1):38767. https://doi.org/10.1038/s41598-025-22626-7

81. Athama A, Srivastava A, Huang S, Wang K, Li Y, Zhai X. Wireless Single-Camera Markerless Motion Capture System for Healthcare Applications. 2025 IEEE International Conference on High Performance Computing and Communications (HPCC). 2025;:694-700. https://doi.org/10.1109/hpcc67675.2025.00106

82. Huang T, Ruan M, Huang S, Fan L, Wu X. Comparison of kinematics and joint moments calculations for lower limbs during gait using markerless and marker-based motion capture. Frontiers in Bioengineering and Biotechnology. 2024;12:1280363. https://doi.org/10.3389/fbioe.2024.1280363

83. Horsak B, Simonlehner M, Quehenberger V, Dumphart B, Wegscheider P, Kranzl A, et al. Validity and reliability of monocular 3D markerless gait analysis in simulated pathological gait: A comparative study with OpenCap. Journal of Biomechanics. 2025;193:112986. https://doi.org/10.1016/j.jbiomech.2025.112986

84. Hou H, Zhao S, Fan Z, Jin W, Zhu J, Ruan L, et al. A Marker-Free Motion Capture System Built on Unsynchronized Cameras. Lecture Notes in Computer Science. 2025;:567-578. https://doi.org/10.1007/978-981-95-2098-5_48

85. Nguyen KX, Zheng L, Hawke AL, Carey RE, Breloff SP, Li K, et al. Deep learning-based estimation of whole-body kinematics from multi-view images. Computer Vision and Image Understanding. 2023;235:103780. https://doi.org/10.1016/j.cviu.2023.103780

86. Li J, Wang Z, Wang C, Su W. GaitFormer: Leveraging dual-stream spatial-temporal Vision Transformer via a single low-cost RGB camera for clinical gait analysis. Knowledge-Based Systems. 2024;295:111810. https://doi.org/10.1016/j.knosys.2024.111810

87. Kumar KS, Jamsrandorj A, Kim J, Mun KR. Prediction of lower limb kinematics from vision-based system using deep learning approaches. 2022 44th Annual International Conference of the IEEE Engineering in Medicine & Biology Society (EMBC). 2022;:177-181. https://doi.org/10.1109/embc48229.2022.9871577

88. Bittner M, Yang WT, Zhang X, Seth A, van Gemert J, van der Helm FCT. Towards Single Camera Human 3D-Kinematics. Sensors. 2023;23(1):341. https://doi.org/10.3390/s23010341

89. Salehi M, Taheri A, Choi S, Kim JH. Evaluation of a markerless motion capture to measure 3D joint kinematics during occupational lifting tasks using mobile devices. Applied Ergonomics. 2026;134:104743. https://doi.org/10.1016/j.apergo.2026.104743

90. Verheul J, Hughes O, Hitchens L, Atherton T, Sauter T, Radwan A, et al. Markerless motion capture for running: validity and reliability of whole-body, joint, and muscle kinematics. Journal of Biomechanics. 2026;195:113133. https://doi.org/10.1016/j.jbiomech.2025.113133

91. Zago M, Luzzago M, Marangoni T, De Cecco M, Tarabini M, Galli M. 3D Tracking of Human Motion Using Visual Skeletonization and Stereoscopic Vision. Frontiers in Bioengineering and Biotechnology. 2020;8:181. https://doi.org/10.3389/fbioe.2020.00181

92. Turner JA, Chaaban CR, Padua DA. Validation of OpenCap: A low-cost markerless motion capture system for lower-extremity kinematics during return-to-sport tasks. Journal of Biomechanics. 2024;171:112200. https://doi.org/10.1016/j.jbiomech.2024.112200

93. Torvinen P, Ruotsalainen KS, Zhao S, Cronin N, Ohtonen O, Linnamo V. Evaluation of 3D Markerless Motion Capture System Accuracy during Skate Skiing on a Treadmill. Bioengineering. 2024;11(2):136. https://doi.org/10.3390/bioengineering11020136

94. Pagnon D, Domalain M, Reveret L. Pose2Sim: An End-to-End Workflow for 3D Markerless Sports Kinematics-Part 2: Accuracy. Sensors. 2022;22(7):2712. https://doi.org/10.3390/s22072712

95. Needham L, Evans M, Cosker DP, Colyer SL. Development, evaluation and application of a novel markerless motion analysis system to understand push-start technique in elite skeleton athletes. PLOS ONE. 2021;16(11):e0259624. https://doi.org/10.1371/journal.pone.0259624

96. Martis P, Kosutska Z, Kranzl A. A Step Forward Understanding Directional Limitations in Markerless Smartphone-Based Gait Analysis: A Pilot Study. Sensors. 2024;24(10):3091. https://doi.org/10.3390/s24103091

97. Needham L, Evans M, Wade L, Cosker DP, McGuigan MP, Bilzon JL, et al. The development and evaluation of a fully automated markerless motion capture workflow. Journal of Biomechanics. 2022;144:111338. https://doi.org/10.1016/j.jbiomech.2022.111338

98. Moro M, Marchesi G, Hesse F, Odone F, Casadio M. Markerless vs. Marker-Based Gait Analysis: A Proof of Concept Study. Sensors. 2022;22(5):2011. https://doi.org/10.3390/s22052011

99. Lim W. Markerless Motion Capture System Based on Webcams Using OpenPose. International Journal of Human Movement and Sports Sciences. 2022;10(5):900-905. https://doi.org/10.13189/saj.2022.100505

100. Lima YL, Collings T, Hall M, Bourne MN, Diamond LE. Validity and reliability of trunk and lower-limb kinematics during squatting, hopping, jumping and side-stepping using OpenCap markerless motion

capture application. Journal of Sports Sciences. 2024;42(19):1847-1858. https://doi.org/10.1080/02640414.2024.2415233

101. Kakavand R, Ahmadi R, Parsaei A, Edwards WB, Komeili A. Comparison of kinematics and kinetics between OpenCap and a marker-based motion capture system in cycling. Computers in Biology and Medicine. 2025;192:110295. https://doi.org/10.1016/j.compbiomed.2025.110295

102. Horsak B, Prock K, Krondorfer P, Siragy T, Simonlehner M, Dumphart B. Inter-trial variability is higher in 3D markerless compared to marker-based motion capture: Implications for data post-processing and analysis. Journal of Biomechanics. 2024;166:112049. https://doi.org/10.1016/j.jbiomech.2024.112049

103. Horsak B, Eichmann A, Lauer K, Prock K, Krondorfer P, Siragy T, et al. Concurrent validity of smartphone-based markerless motion capturing to quantify lower-limb joint kinematics in healthy and pathological gait. Journal of Biomechanics. 2023;159:111801. https://doi.org/10.1016/j.jbiomech.2023.111801

104. Harrison SM, Cohen RCZ, Starkey SC, Greenwood J, Cheong E, Nguyen K, et al. Evaluation of the Ergomechanic Markerless Motion Capture System for Lower Body Kinematics During Standing, Squatting, and Walking. Journal of Biomechanical Engineering. 2025;147(12):121003. https://doi.org/10.1115/1.4069821

105. Emmerson J, Needham L, Williams S, Colyer S. Evaluation of a markerless motion capture system for measuring mechanical work during tennis strokes. Journal of Sports Sciences. 2025;44(10):1348-1360. https://doi.org/10.1080/02640414.2025.2555096

106. D'Haene M, Chorin F, Colson SS, Guérin O, Zory R, Piche E. Validation of a 3D Markerless Motion Capture Tool Using Multiple Pose and Depth Estimations for Quantitative Gait Analysis. Sensors. 2024;24(22):7105. https://doi.org/10.3390/s24227105

107. de Borba EF, Storniolo JL, Cerfoglio S, Capodaglio P, Cimolin V, Peyré-Tartaruga LA, et al. Effect of Walking Speed on the Reliability of a Smartphone-Based Markerless Gait Analysis System. Sensors. 2025;25(20):6474. https://doi.org/10.3390/s25206474

108. D'Antonio E, Taborri J, Palermo E, Rossi S, Patane F. A markerless system for gait analysis based on OpenPose library. 2020 IEEE International Instrumentation and Measurement Technology Conference (I2MTC). 2020;:1-6. https://doi.org/10.1109/i2mtc43012.2020.9128918

109. D'Antonio E, Taborri J, Mileti I, Rossi S, Patane F. Validation of a 3D Markerless System for Gait Analysis based on OpenPose and Two RGB Webcams. IEEE Sensors Journal. 2021;21(15):17064-17075. https://doi.org/10.1109/jsen.2021.3081188

110. D'Antonio E, Taborri J, Palermo E, Rossi S, Patane F. Characterization of a low-cost markerless system for 3D gait analysis. 2020 IEEE International Symposium on Medical Measurements and Applications (MeMeA). 2020;:1-6. https://doi.org/10.1109/memea49120.2020.9137236

111. Bertozzi F, Brunetti C, Maver P, Palombi M, Santini M, Galli M, et al. Concurrent validity of IMU and phone-based markerless systems for lower-limb kinematics during cognitively-challenging landing tasks. Journal of Biomechanics. 2025;191:112883. https://doi.org/10.1016/j.jbiomech.2025.112883

112. Bak SY, Ahn J, Choi H, Lee S, Lim W, Kim HD. Validity of Markerless Motion Capture System and Its Correlation with Physical Characteristics for Hip Range of Motion Measurement: A Pilot Study. Sensors and Materials. 2025;37(3):965. https://doi.org/10.18494/sam5489

113. Bae K, Lee S, Bak SY, Kim HS, Ha Y, You JH. Concurrent validity and test reliability of the deep learning markerless motion capture system during the overhead squat. Scientific Reports. 2024;14(1). https://doi.org/10.1038/s41598-024-79707-2

114. Ahn J, Choi H, Lee H, Kim SW, Lee J, Kim HD. Simultaneous Validity and Intra-Test Reliability of Joint Angle Measurement through Novel Multi-RGB Sensor-Based Three-Joint-Continuous-Motion Analysis: A Pilot Study. Applied Sciences. 2023;14(1):73. https://doi.org/10.3390/app14010073

115. Verheul J, Robinson MA, Burton S. Jumping towards field-based ground reaction force estimation and assessment with OpenCap. Journal of Biomechanics. 2024;166:112044. https://doi.org/10.1016/j.jbiomech.2024.112044

116. Wang J, Xu W, Wu Z, Zhang H, Wang B, Zhou Z, et al. Evaluation of a smartphone-based markerless system to measure lower-limb kinematics in patients with knee osteoarthritis. Journal of Biomechanics. 2025;181:112529. https://doi.org/10.1016/j.jbiomech.2025.112529

117. Bousigues S, Naaim A, Robert T, Muller A, Dumas R. The effects of markerless inconsistencies are at least as large as the effects of the marker-based soft tissue artefact. Journal of Biomechanics. 2025;182:112566. https://doi.org/10.1016/j.jbiomech.2025.112566

118. Templin T, Riehm CD, Eliason T, Hulburt TC, Kwak ST, Medjaouri O, et al. Evaluation of drop vertical jump kinematics and kinetics using 3D markerless motion capture in a large cohort. Frontiers in Bioengineering and Biotechnology. 2024;12. https://doi.org/10.3389/fbioe.2024.1426677

119. Ho MY, Kuo MC, Chen CS, Wu RM, Chuang CC, Shih CS, et al. Pathological Gait Analysis With an Open-Source Cloud-Enabled Platform Empowered by Semi-Supervised Learning-PathoOpenGait. IEEE Journal of Biomedical and Health Informatics. 2024;28(2):1066-1077. https://doi.org/10.1109/jbhi.2023.3340716

120. Chen L, Zheng Y, Gong Z, Wang D. Gait anomaly detection based on video-derived 3D pose estimation. Medical & Biological Engineering & Computing. 2025;63(9):2651-2663. https://doi.org/10.1007/s11517-025-03339-5

121. Fukushima T, Blauberger P, Guedes Russomanno T, Lames M. The potential of human pose estimation for motion capture in sports: a validation study. Sports Engineering. 2024;27(1). https://doi.org/10.1007/s12283-024-00460-w

122. Svetek A, Morgan K, Burland J, Glaviano NR. Validation of OpenCap on lower extremity kinematics during functional tasks. Journal of Biomechanics. 2025;183:112602. https://doi.org/10.1016/j.jbiomech.2025.112602

123. Hu B, Wang J, Xu W, Li T, Nie Y, Li K. Dual-Camera Markerless Motion Capture System for Precise Lower-Limb Kinematic Analysis in Osteoarthritis. Annals of Biomedical Engineering. 2025;53(11):2949-2965. https://doi.org/10.1007/s10439-025-03859-z

124. Vafadar S, Skalli W, Bonnet-Lebrun A, Assi A, Gajny L. Assessment of a novel deep learning-based marker-less motion capture system for gait study. Gait & Posture. 2022;94:138-143. https://doi.org/10.1016/j.gaitpost.2022.03.008

125. Yeom S, Jeong B, Lee J, Hyung SK, Kim H, Chang E, et al. Validation of Markerless Motion Capture System (OpenCap) for Overhead Squat Kinematics. Measurement in Physical Education and Exercise Science. 2025;30(1):49-62. https://doi.org/10.1080/1091367x.2025.2512361

126. Todesca D, Fabiocchi D, Farías Fuentes J, Giulietti N, Carnevale M, Giberti H. Comparison of Real-Time Marker-Less and Optoelectronic 3D Human Pose Estimation Systems for Cyclist Pose Analysis. 2025 IEEE International Conference on Metrology for eXtended Reality, Artificial Intelligence and Neural Engineering (MetroXRAINE). 2025:417-422. https://doi.org/10.1109/metroxraine66377.2025.11340187

127. Zhang C, Li Y, Ye W, Huang G. Human Kinematics Analysis by Markerless Vision Based on OpenSim. 2024 IEEE 5th International Conference on Pattern Recognition and Machine Learning (PRML). 2024:443-450. https://doi.org/10.1109/prml62565.2024.10779779

128. Liang Y, Qi S, Xu T, Hu Y. 3D Gait Analysis for the Elderly Mobility Based on Multiple RGB Cameras. 2023 29th International Conference on Mechatronics and Machine Vision in Practice (M2VIP). 2023;:1-5. https://doi.org/10.1109/m2vip58386.2023.10413409

129. Zhang L, Sidarta A, Wu TL, Jatesiktat P, Wang H, Li L, et al. Towards Clinical Application of Enhanced Timed Up and Go with Markerless Motion Capture and Machine Learning for Balance and Gait Assessment. IEEE Journal of Biomedical and Health Informatics. 2025:1-9. https://doi.org/10.1109/JBHI.2025.3543095

130. Shah EEZ, Mustansar Z. Biomechanics of Knee Kinematics During Sit-Ups Using MoCap Technology. 2025 27th International Multitopic Conference (INMIC). 2025;:1-4. https://doi.org/10.1109/inmic65900.2025.11348465

131. McGinley JL, Baker R, Wolfe R, Morris ME. The reliability of three-dimensional kinematic gait measurements: A systematic review. Gait Posture. 2009;29(3):360-369. https://doi.org/10.1016/j.gaitpost.2008.09.003

132. Soesilo TH, Gunaratne PN, Tamura H. Development of a Real-Time Multi-Person 3D Keypoint Detection System Using Stereoscopic Cameras and RTMPose. In: Proceedings of the 2025 International Conference on Artificial Life and Robotics (ICAROB2025). Oita, Japan: ALife Robotics; 2025. p. 593-596.

133. Wang Y, Wang Z, Liu L, Daniilidis K. TRAM: Global Trajectory and Motion of 3D Humans from in-the-wild Videos. arXiv:2403.17346 [cs.CV]. 2024.

134. Bermuth D, Poeppel A, Reif W. RapidPoseTriangulation: Multi-view Multi-person Whole-body Human Pose Triangulation in a Millisecond. arXiv:2503.21692 [cs.CV]. 2025.

135. Janampa S, Pattichis M. DETRPose: Real-time end-to-end transformer model for multi-person pose estimation. arXiv:2506.13027 [cs.CV]. 2025.

136. Mehraban S, Iaboni A, Taati B. FastHMR: Accelerating Human Mesh Recovery via Token and Layer Merging with Diffusion Decoding. arXiv:2510.10868 [cs.CV]. 2025.

137. Cho H, Choi G, Choi J. AJAHR: Amputated Joint Aware 3D Human Mesh Recovery. arXiv:2509.19939 [cs.CV]. 2025.

138. Zheng Z, Yang L, Pan J, Zhu H. Mamba-Driven Topology Fusion for Monocular 3D Human Pose Estimation. arXiv:2505.20611 [cs.CV]. 2025.

139. Bright J, Chen Y, Zelek JS. DreamPose3D: Hallucinative Diffusion with Prompt Learning for 3D Human Pose Estimation. arXiv:2511.09502 [cs.CV]. 2025.

140. Han B, Huang Y, Gao P. HyperDiff: Hypergraph Guided Diffusion Model for 3D Human Pose Estimation. arXiv:2508.14431 [cs.CV]. 2025.

141. Armstrong K, Rodrigues A, Willmott AP, Zhang L, Ye X. Validation of Human Pose Estimation and Human Mesh Recovery for Extracting Clinically Relevant Motion Data from Videos. arXiv:2503.14760 [cs.CV]. 2025.

142. Pemasiri A, Goan E, Lichtwark G, Schuster R, Kelly L, Fookes C. Biomechanically Accurate Gait Analysis: A 3D Human Reconstruction Framework for Markerless Estimation of Gait Parameters. arXiv:2603.02499 [cs.CV]. 2026.

143. Guo C, L'Erario G, Romualdi G, Leonori M, Lorenzini M, Ajoudani A, et al. Physics-Informed Learning for Human Whole-Body Kinematics Prediction via Sparse IMUs. arXiv:2509.25704 [cs.CV]. 2025.

144. Kinfu KA, Vidal R. Efficient Vision Transformer for Human Pose Estimation via Patch Selection. arXiv:2306.04225 [cs.CV]. 2023.

145. Fang H, Cai J, Wang X, Yang W. Beyond Static Frames: Temporal Aggregate-and-Restore Vision Transformer for Human Pose Estimation. arXiv:2603.05929 [cs.CV]. 2026.

146. Yang R, Kennedy A, Cotton RJ. BiomechGPT: Towards a Biomechanically Fluent Multimodal Foundation Model for Clinically Relevant Motion Tasks. arXiv:2505.18465 [cs.CV]. 2025.

147. Cotton RJ, Leonard T. BiomechAgent: AI-Assisted Biomechanical Analysis Through Code-Generating Agents. arXiv:2602.06975 [cs.CV]. 2026.

148. Katsu J, et al. GRF-MV: Ground Reaction Force Estimation from Monocular Video. In: BMVC 2024 Workshop (ANIMA). 2024.

149. Han X, Senderling B, To S, Kumar D, Whiting E, Saito J. GroundLink: A Dataset Unifying Human Body Movement and Ground Reaction Dynamics. arXiv:2310.03930 [cs.CV]. 2023.

150. Hossain MSB, Choi H, Guo Z, Yoo S, Song MK, Shin H, et al. Knowledge transfer-driven estimation of knee moments and ground reaction forces from smartphone videos via temporal-spatial modeling of augmented joint kinematics. PLOS One. 2025;20(11):e0335257. https://doi.org/10.1371/journal.pone.0335257

151. Jiang T, Lu P, Zhang L, Ma N, Han R, Lyu C, et al. RTMPose: real-time multi-person pose estimation based on MMPose. arXiv:2303.07399. 2023.

152. Xu Y, Zhang J, Zhang Q, Tao D. ViTPose: simple vision transformer baselines for human pose estimation. Adv Neural Inf Process Syst (NeurIPS). 2022. arXiv:2204.12484.

153. Khirodkar R, Bagautdinov T, Martinez J, Zhaoen S, James A, Selednik P, et al. Sapiens: Foundation for Human Vision Models. arXiv:2408.12569 [cs.CV]. 2024.

154. Huang Y, Liu J, Xian K, Qiu RC. PoseMamba: Monocular 3D Human Pose Estimation with Bidirectional Global-Local Spatio-Temporal State Space Model. arXiv:2408.03540 [cs.CV]. 2024.

155. Yang X, Kukreja D, Pinkus D, Sagar A, Fan T, Park J, et al. SAM 3D Body: Robust Full-Body Human Mesh Recovery. arXiv:2602.15989 [cs.CV]. 2026.

156. Shin S, Kim J, Halilaj E, Black MJ. WHAM: Reconstructing World-grounded Humans with Accurate 3D Motion. arXiv:2312.07531 [cs.CV]. 2023.

157. Shen Z, Pi H, Xia Y, Cen Z, Peng S, Hu Z, et al. World-Grounded Human Motion Recovery via Gravity-View Coordinates. arXiv:2409.06662 [cs.CV]. 2024.

158. Sárándi I, Pons-Moll G. Neural Localizer Fields for Continuous 3D Human Pose and Shape Estimation. arXiv:2407.07532 [cs.CV]. 2024.

159. Gozlan Y, Falisse A, Uhlrich S, Gatti A, Black M, Chaudhari A. OpenCapBench: A Benchmark to Bridge Pose Estimation and Biomechanics. arXiv:2406.09788 [cs.CV]. 2024.

160. Unger T, Sal Moslehian A, Peiffer JD, Ullrich J, Gassert R, Lambercy O, et al. Differentiable Biomechanics for Markerless Motion Capture in Upper Limb Stroke Rehabilitation: A Comparison with Optical Motion Capture. arXiv:2411.14992 [cs.CV]. 2024.

161. Voss H, Kopp S. JAX-IK: Real-Time Inverse Kinematics for Generating Multi-Constrained Movements of Virtual Human Characters. arXiv:2507.00792 [cs.CV]. 2025.

162. Tolpin D, Kagarlitsky S. Fast Neural Inverse Kinematics on Human Body Motions. arXiv:2506.17996 [cs.CV]. 2025.

163. Tan T, Van Wouwe T, Werling K, Liu CK, Delp S, Hicks J, et al. GaitDynamics: a generative foundation model for analyzing human walking and running. Nat Biomed Eng. 2026. https://doi.org/10.1038/s41551-025-01565-8

164. Wehrbein T, Rudolph M, Rosenhahn B, Wandt B. Probabilistic Monocular 3D Human Pose Estimation with Normalizing Flows. In: Proceedings of the IEEE/CVF International Conference on Computer Vision (ICCV). 2021. arXiv:2107.13788 [cs.CV].

165. Fritz S, Lusardi M. White Paper: "Walking Speed: the Sixth Vital Sign". J Geriatr Phys Ther. 2009;32(2):2-5. https://doi.org/10.1519/00139143-200932020-00002

166. Yang T, He S, Jing H, Yang J, Liu Z, Zou C, et al. Fast SAM 3D Body: Accelerating SAM 3D Body for Real-Time Full-Body Human Mesh Recovery. arXiv:2603.15603 [cs.CV]. 2026.

167. Elbatel M, Wang A, Liu K, Mouheb K, Almar-Munoz E, Lin L, et al. MedSapiens: Taking a Pose to Rethink Medical Imaging Landmark Detection. arXiv:2511.04255 [cs.CV]. 2025.

168. Werling K, Bianco NA, Raitor M, Stingel J, Hicks JL, Collins SH, et al. AddBiomechanics: Automating model scaling, inverse kinematics, and inverse dynamics from human motion data through sequential optimization. PLOS ONE. 2023;18(11):e0295152. https://doi.org/10.1371/journal.pone.0295152

169. Cuevas-Velasquez H, Yiannakidis A, Shin S, Becherini G, Höschle M, Tesch J, et al. MAMMA: Markerless & Automatic Multi-Person Motion Action Capture. arXiv:2506.13040 [cs.CV]. 2025.

170. Peiffer JD, Shah K, Djuraskovic I, Anarwala S, Abdou K, Patel R, et al. Portable Biomechanics Laboratory: Clinically Accessible Movement Analysis from a Handheld Smartphone. arXiv:2507.08268 [cs.CV]. 2025.

171. Li J, Bian S, Xu C, Chen Z, Yang L, Lu C. HybrIK-X: Hybrid Analytical-Neural Inverse Kinematics for Whole-body Mesh Recovery. arXiv:2304.05690 [cs.CV]. 2023.

172. Wang Y, Sun Y, Patel P, Daniilidis K, Black MJ, Kocabas M. PromptHMR: Promptable Human Mesh Recovery. arXiv:2504.06397 [cs.CV]. 2025.

173. Fiche G, Weinzaepfel P, Bregier R, Baradel F. Multi-HMR 2: Multi-Person Camera-Centric Human Detection, Mesh Recovery and Tracking. arXiv:2606.14841 [cs.CV]. 2026.

174. Wang R, Xu S, Dong Y, Deng Y, Xiang J, Lv Z, et al. MoGe-2: Accurate Monocular Geometry with Metric Scale and Sharp Details. arXiv:2507.02546 [cs.CV]. 2025.

175. Piccinelli L, Sakaridis C, Yang Y-H, Segu M, Li S, Abbeloos W, Van Gool L. UniDepthV2: Universal Monocular Metric Depth Estimation Made Simpler. arXiv:2502.20110 [cs.CV]. 2025.

176. Lee SE, Nishino K, Nobuhara S. SteerPose: Simultaneous Extrinsic Camera Calibration and Matching from Articulation. arXiv:2506.01691 [cs.CV]. 2025.

177. Patel P, Black MJ. CameraHMR: Aligning People with Perspective. arXiv:2411.08128 [cs.CV]. 2024.

## Additional files

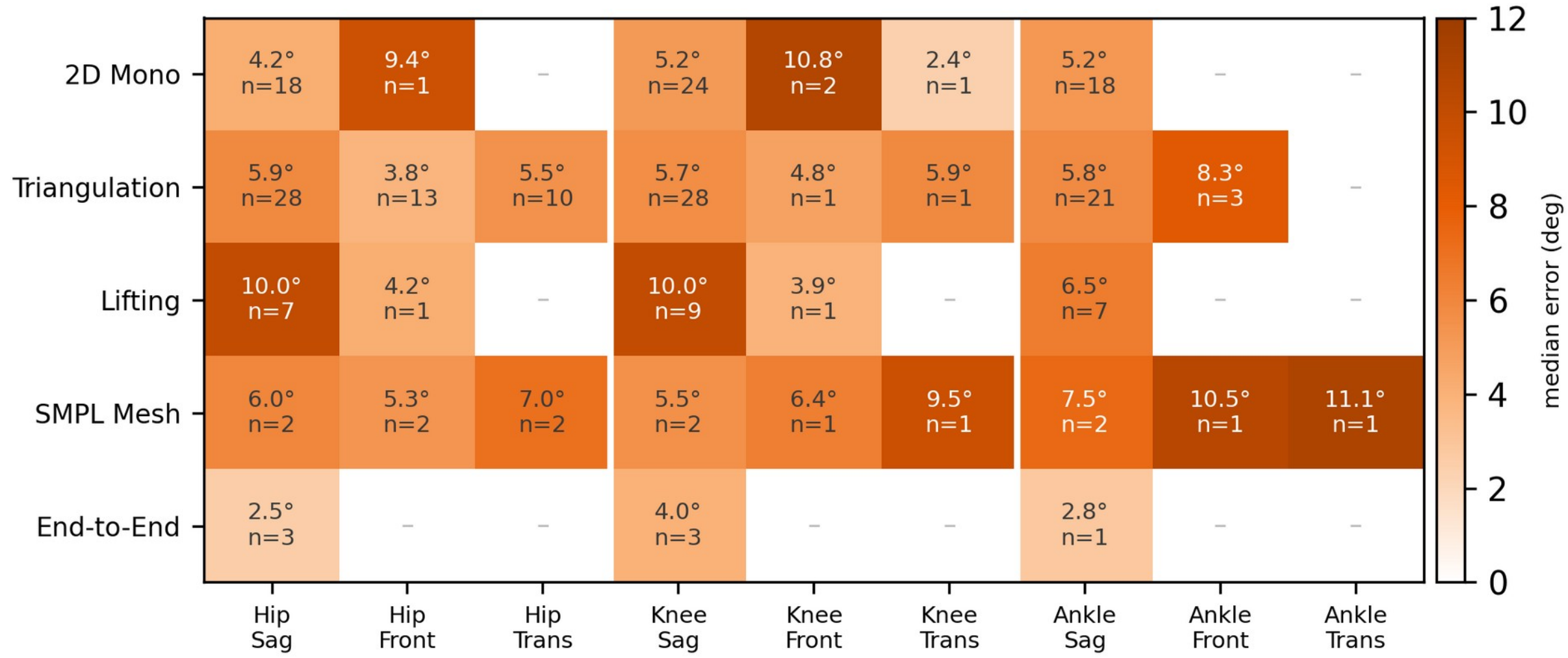

Additional file 1: PRISMA-ScR reporting checklist.

Additional file 2: Appendix A, full search strings per database.

Additional file 3: complete extraction table (117 studies, about 90 fields), machine-readable.

Additional file 4: median lower-limb error by architecture, joint and plane (figure).

*Additional file 5: full curated computer-vision corpus (MICRO; 572 methods across the pipeline stages), machine-readable. Methods featured in the main-text state of the art (the toolbox in Fig. 7 and those discussed in the accompanying text) are flagged and listed first; the corpus is a deliberately non-exhaustive horizon scan.*

| **Method [ref]** | **Pipeline stage** | **What it adds** | **Maturity** |
|---|---|---|---|
| MoGe-2 [174] | Input / scene geometry | Single-image metric depth and 3D geometry (scene scale) | Emerging |
| UniDepth v2 [175] | Input / scene geometry | Metric monocular depth | Emerging |
| SteerPose [176] | Input / calibration | Target-free multi-camera self-calibration from the moving person | Emerging |
| TRAM [133] | Input / world grounding | Camera-motion and gravity grounding in world coordinates | Emerging |
| CameraHMR [177] | Input / world grounding | Learned field-of-view for metric placement | Emerging |
| RTMPose [151] | 2D pose | Real-time 2D keypoint backbone | Established (CV) |
| ViTPose [152] | 2D pose | Transformer 2D keypoint backbone | Established (CV) |
| Sapiens [153] | 2D pose / features | Human-centric foundation backbone | Emerging |
| MotionBERT [10] | 3D lifting | Monocular 2D-to-3D pose lifting | Established (CV) |
| PoseMamba [154] | 3D lifting | State-space 2D-to-3D lifting | Emerging |
| SAM 3D Body [155] | Mesh recovery | Foundation parametric-mesh (SMPL) recovery | Emerging |
| WHAM [156] | Mesh recovery | World-grounded human mesh from video | Emerging |
| GVHMR [157] | Mesh recovery | Gravity-view human mesh recovery | Emerging |
| PromptHMR [172] | Mesh recovery | Promptable human mesh recovery | Emerging |
| Neural Localizer Fields [158] | Mesh to anatomy | Queries arbitrary anatomical points on the mesh | Emerging |
| HybrIK-X [171] | Mesh to angles | Analytic-neural inverse kinematics from mesh | Emerging |
| Multi-HMR 2 [173] | Mesh recovery | Multi-person camera-centric SMPL-X with tracking | Emerging |
| MAMMA [169] | Multi-view mesh | Model-space multi-view mesh fitting | Emerging |
| Pose2Sim [8] | Triangulation to OpenSim | Classical multi-view triangulation to a musculoskeletal solve | Established |
| SynthPose / OpenCapBench [159] | Biomechanical bridge | Virtual-marker augmentation with joint-angle evaluation | Emerging |
| Differentiable | Biomechanical bridge | Differentiable, model-based | Emerging |
| Biomechanics [160] | | inverse kinematics | |
| JAX-IK [161] | Biomechanical bridge | Real-time differentiable inverse kinematics | Emerging |
| Fast Neural IK [162] | Biomechanical bridge | Learned (neural) inverse kinematics | Emerging |
| AddBiomechanics [168] | Biomechanical bridge | Offline scaling, inverse kinematics and inverse dynamics | Emerging |
| GaitDynamics [163] | Kinetics | Generative estimation of ground-reaction forces from kinematics | Emerging |
| BiomechGPT [146] | Analysis layer | Multimodal biomechanics foundation model for motion queries | Emerging |
| Wehrbein et al. [164] | Uncertainty | Distributional pose estimation via normalizing flows | Emerging |
| BioPose [11] | Near-complete pipeline | Mesh recovery plus neural inverse-kinematics solve | Emerging |
| OpenCap Monocular [12] | Near-complete pipeline | WHAM mesh + physics refinement + OpenSim (with dynamics) | Emerging |
| Portable Biomechanics Laboratory [170] | Near-complete pipeline | Handheld video to OpenSim with clinically acceptable agreement | Emerging |

Additional file 6: MICRO corpus size by pipeline-step block (figure).

Additional file 7: MICRO corpus construction and reproducible arXiv screening protocol.